\documentclass[conference]{IEEEtran}

\IEEEoverridecommandlockouts
\usepackage{cite}
\usepackage{amsmath,amssymb,amsfonts}
\usepackage{algorithmic}
\usepackage{graphicx}
\usepackage{textcomp}
\usepackage{xcolor}
\usepackage{amsmath}
\usepackage{amsfonts}
\usepackage{multirow}
\usepackage{subfigure}
\usepackage{algorithm,float}
\usepackage{adjustbox}
\usepackage{booktabs}
\usepackage{makecell}
\usepackage{hyperref}
\newtheorem{prop}{Proposition}
\def\BibTeX{{\rm B\kern-.05em{\sc i\kern-.025em b}\kern-.08em
    T\kern-.1667em\lower.7ex\hbox{E}\kern-.125emX}}
\begin{document}
\newcommand\autoac{AutoAC}
\title{AutoAC: Towards Automated Attribute
Completion for Heterogeneous Graph Neural
Network (Extended Version)}
\author{\IEEEauthorblockN{Guanghui Zhu,
Zhennan Zhu,
Wenjie Wang,
Zhuoer Xu,
Chunfeng Yuan,
Yihua Huang
}
\IEEEauthorblockA{\it State Key Laboratory for Novel Software Technology, Nanjing University, Nanjing, China \\
Department of Computer
Science and Technology, Nanjing
University, Nanjing, China \\
\{zhuzhennan, wenjie.wang, zhuoer.xu\}@smail.nju.edu.cn, \{zgh, cfyuan, yhuang\}@nju.edu.cn}
}

\maketitle

\begin{abstract}
 Many real-world data can be modeled as heterogeneous graphs that contain multiple types of nodes and edges.
  Meanwhile, due to excellent performance, heterogeneous graph neural networks (GNNs) have received more and more attention.
  However, the existing work mainly focuses on the design of novel GNN models, while ignoring another important issue that also has a large impact on the model performance, namely the missing attributes of some node types.
  The handcrafted attribute completion requires huge expert experience and domain knowledge.
  Also, considering the differences in semantic characteristics between
  nodes, the attribute completion should be fine-grained, i.e., the attribute completion operation should be node-specific.  
  Moreover, to improve the performance of the downstream graph learning task, attribute completion and the training of the heterogeneous GNN should be jointly optimized rather than viewed as two separate processes.
  To address the above challenges, we propose a differentiable attribute completion framework called \autoac{} for automated completion operation search in heterogeneous GNNs.
  We first propose an expressive completion operation search space, including topology-dependent and topology-independent completion operations.
  Then, we propose a continuous relaxation schema and further propose a differentiable completion algorithm where the completion operation search is formulated as a bi-level joint optimization problem.
  To improve the search efficiency, we leverage two optimization techniques: discrete constraints and auxiliary unsupervised graph node clustering.
  Extensive experimental results on real-world datasets reveal that \autoac{} outperforms the SOTA handcrafted heterogeneous GNNs and the existing attribute completion method.
\end{abstract}

\begin{IEEEkeywords}
heterogeneous graph, graph neural network, attribute completion, differentiable search
\end{IEEEkeywords}
\section{Introduction}
\begin{figure*}
    \centering
    \subfigure[]{
    \includegraphics[width=0.23\textwidth]{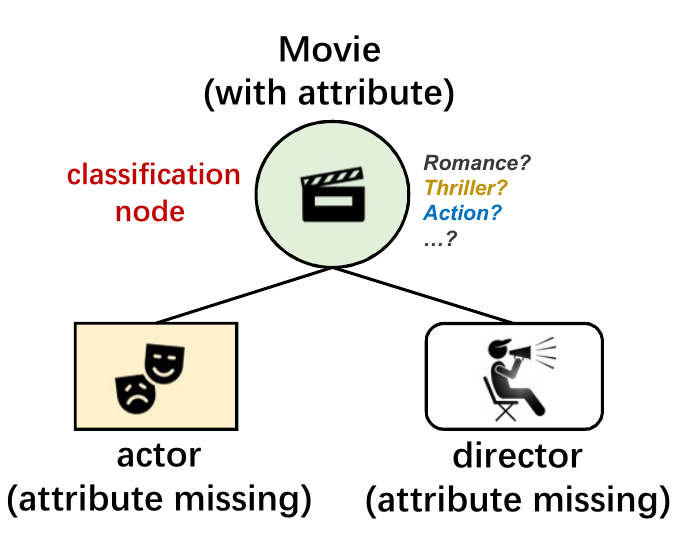}
    }
    \subfigure[]{
    \includegraphics[width=0.60\textwidth]{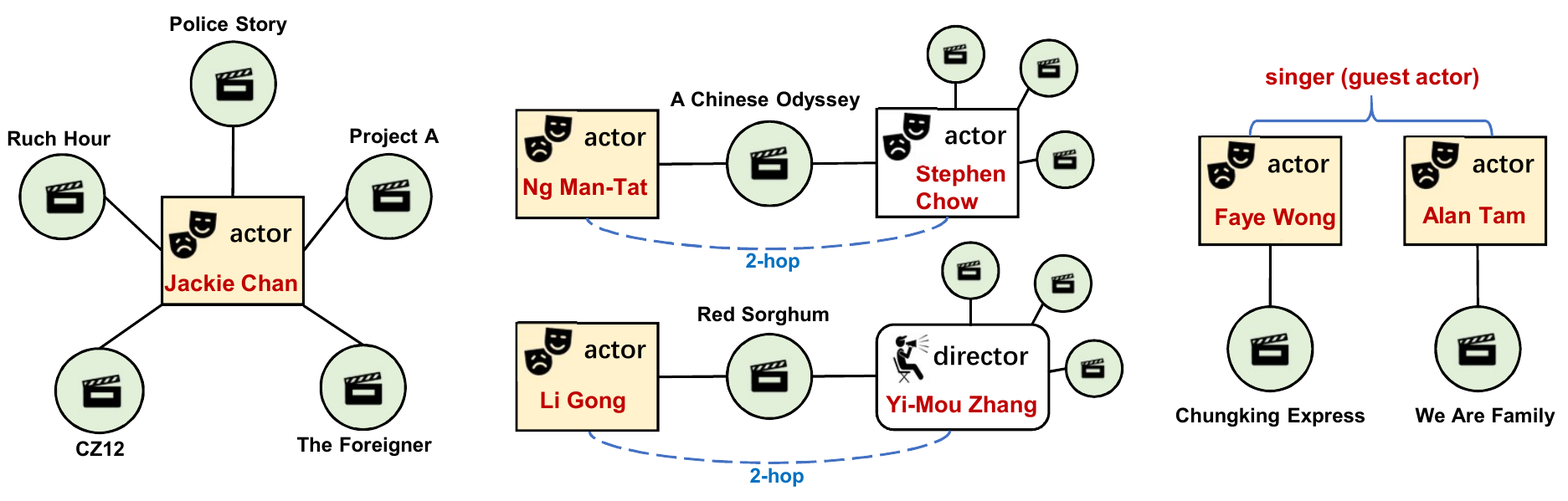}
    }
    \vspace{-1ex}
    \caption{(a) Example of heterogeneous graphs with incomplete attributes, i.e., the IMDB dataset. (b) Different attribute completion operations for the actor node, i.e., local attribute aggregation, message-passing based multi-hop attribute aggregation, and one-hot representation.}
    \label{example}
    \vspace{-3ex}
\end{figure*}

Graph-structured data are ubiquitous, such as social networks~\cite{hamilton2017inductive}, scholar networks~\cite{sen2008collective}, biochemical networks~\cite{zitnik2017predicting}, and knowledge graphs~\cite{ji2021survey}. 
Meanwhile, many real-world graph data are heterogeneous~\cite{sun2013mining}.
Unlike the homogeneous graph with only one node type and one edge type, the heterogeneous graph~\cite{shi2016survey} consists of multiple types of nodes and edges associated with attributes in different feature spaces. 
For example, the IMDB dataset is a typical heterogeneous graph, which contains three node types (movie, actor, director) and two edge types (movie-actor, movie-director), as shown in Figure~\ref{example}(a). 
Due to containing rich information and semantics, heterogeneous graphs have drawn more and more attention.

Recently, graph neural networks (GNNs)~\cite{wu2020comprehensive, kipf2017gcn} have demonstrated powerful representation learning ability on graph-structured data~\cite{medical2021sigmod}. 
Meanwhile, many heterogeneous GNNs (HGNNs) have been proposed for heterogeneous graphs~\cite{wang2019heterogeneous}\cite{yun2019graph}\cite{zhu2019relation}\cite{zhang2019heterogeneous}\cite{fu2020magnn}\cite{hu2020heterogeneous}\cite{hong2020attention}\cite{lv2021we}. 
However, the existing work on heterogeneous graphs mainly focuses on the construction of novel GNN models, while ignoring another important issue that also has a large impact on the model performance, namely the attributes of some types of nodes are missing~\cite{jin2021heterogeneous}. 
Missing node attributes is a common problem because collecting the attributes of all nodes is prohibitively expensive or even impossible due to privacy concerns. 
Since the attributes of all nodes are required in the GNN-based heterogeneous models, some handcrafted ways are employed to deal with the problem of missing attributes. 
For example, the missing attribute vector can be the sum or the mean of directly connected nodes’ attribute vectors. 
Besides, the one-hot representations of a certain node type can also be used to replace the missing attributes. 
However, the handcrafted ways require huge expert experience and domain knowledge. 
Also, the topological relationships in the graph are not taken into account. 
Recently, an attention-based method \cite{jin2021heterogeneous} was proposed to complete each no-attribute node by weighted aggregation of the attributes from the directly neighboring attributed nodes. 
Such an attribute completion method only considers the attributes of 1-hop neighbors without exploiting the attributes of higher-order neighbors.

Moreover, existing attribute completion methods are all coarse-grained. 
That is, for a specific node type without attributes, they adopt the same attribute completion operation for all nodes without considering the differences in semantic characteristics between nodes. 
In practice, fine-grained attribute completion is more reasonable. 
The attribute completion operations for the nodes with different semantics should be different. 
Take the IMDB dataset as an example. 
The target type of nodes (i.e., movie nodes) has attributes, and the other types of nodes (i.e., actor nodes and director nodes) have no attributes. 
As shown in Figure~\ref{example}(b), there exist three attribute completion operations, including 1) For actors (e.g. Jackie Chan) who are involved in movies that mostly belong to the same genre (Kung Fu movies), average attribute aggregation of local (i.e., 1-hop) neighboring nodes should be used. 2) 
For actors who have strong collaborative relationships with other actors and directors, the message-passing based multi-hop attribute aggregation is more suitable. 3) For guest actors without representative movies, we can directly use the simple one-hot encoding to complete attributes.

\textcolor{ black}{For the IMDB dataset, the number of actor nodes that have no attributes is 6124. Manually differentiating the semantic characteristics of all no-attribute nodes and then selecting the most suitable completion operations according to semantic characteristics is infeasible.}
%
Thus, an automated attribute completion method that can search the optimal completion operations efficiently is required. 
Moreover, to improve the performance of the downstream graph learning task, the automated attribute completion and the training of the heterogeneous GNN should be jointly optimized rather than viewed as two separate processes.

To address the above challenges, we propose a differentiable attribute completion framework called \autoac{}\footnote{\autoac{} is available at https://github.com/PasaLab/AutoAC} for automated completion operation search in heterogeneous GNNs. 
\autoac{} is a generic framework since it can integrate different heterogeneous GNNs flexibly. 
By revisiting the existing attribute completion methods, we first propose an expressive completion operation search space, including topology-dependent and topology-independent completion operations. %
Instead of searching over the discrete space (\textcolor{ black}{i.e., candidate completion operations for each no-attribute node}), we propose a continuous relaxation scheme by placing a weighted mixture of candidate completion choices, which turns the search task into an optimization problem regarding the weights of choices (i.e., completion parameters). 
%
\textcolor{ black}{Thus, due to the continuous search space, the search process becomes differentiable and we can perform completion operation search via gradient descent.}

To further improve the search efficiency, we formulate the search of attribute completion operations and the training of GNN as a constrained bi-level joint optimization problem.
Specifically, we keep the search space continuous in the optimization process of completion parameters (i.e., upper-level optimization) but enforce attribute completion choices being discrete in the optimization process of weights in the heterogeneous GNN (i.e., lower-level optimization). 
In this way, there is only one activated completion operation for each no-attribute node during the training of GNN, removing the need to perform all candidate completion operations.
Inspired by NASP~\cite{yao2020efficient}, we employ proximal iteration to solve the constrained optimization problem efficiently.

Finally, to reduce the dimension of the attribute completion parameters, we further leverage an auxiliary unsupervised graph node clustering task with the spectral modularity function during the process of GNN training.

To summarize, the main contributions of this paper can be
highlighted as follows:
\begin{itemize}
\item We are the first, to the best of our knowledge, to model the attribute completion problem as an automated search problem for the optimal completion operation of each no-attribute node. 
\item We propose an expressive completion operation search space and further propose a differentiable attribute completion framework where the completion operation search is formulated as a bi-level joint optimization problem.
\item To improve search efficiency, we enforce discrete constraints on completion parameters in the training of heterogeneous GNN. Moreover, we leverage an auxiliary unsupervised graph node clustering task to reduce the dimension of the attribute completion parameters.
\item Extensive experimental results on real-world datasets reveal that \autoac{} is effective to boost the performance of heterogeneous GNNs and outperforms the SOTA attribute completion method in terms of performance and efficiency.
\end{itemize}
\section{Related Work}
\subsection{Heterogeneous Graph Neural Network}
Graph neural network~\cite{kipf2017gcn}\cite{velickovic2017graph}\cite{hamilton2017inductive}\cite{10.1145/3514221.3517872}\cite{10.1145/3318464.3389706}\cite{medical2021sigmod} aims to extend neural networks to graphs. 
Since heterogeneous graphs are more common in the real world~\cite{sun2013mining}, heterogeneous GNNs have been proposed recently. 
Part of the work is based on meta-paths. 
HAN~\cite{wang2019heterogeneous} leverages the semantics of meta-paths and uses hierarchical attention to aggregate neighbors.
MAGNN~\cite{fu2020magnn} utilizes RotatE~\cite{sun2018rotate} to encode intermediate nodes along each meta-path and mix multiple meta-paths using hierarchical attention. 
Another part of the work chooses to extract rich semantic information in heterogeneous graphs.
GTN~\cite{yun2019graph} learns a soft selection of edge types and composite relations for generating useful multi-hop connections.
HetGNN~\cite{zhang2019heterogeneous} uses Bi-LSTM to aggregate node features for each type and among types. 
As the state-of-the-art model, SimpleHGN~\cite{lv2021we} revisits existing methods and proposes a simple framework using learnable edge-type embedding and residual connections for both nodes and edges.
\textcolor{ black}{Recently, AS-GCN~\cite{yu2021gcn} employs the heterogeneous
GNN to mine the semantics for text-rich networks.}

Different from the above methods, HGNN-AC~\cite{jin2021heterogeneous} notices that most of the nodes in the real heterogeneous graph have missing attributes, which could cause great harm to the performance of heterogeneous models, and proposes an attention-based attribute completion method.
However, HGNN-AC needs to get node embeddings based on network topology using metapath2vec~\cite{dong2017metapath2vec}, which is a time-consuming process.
Moreover, the attribute completion in HGNN-AC is coarse-grained and supports only one completion operation for all no-attribute nodes. 
\textcolor{ black}{
HGCA~\cite{9724614} unifies attribute completion and representation learning in an unsupervised heterogeneous network.
MRAP~\cite{bayram2021node} performs node attribute competition in knowledge graphs with multi-relational propagation.
}

\subsection{Neural Architecture Search (NAS)}
NAS~\cite{elsken2019neural} that designs effective neural architectures automatically has received more attention. 
%
The core components of NAS contain search space, search algorithm, and performance estimation strategy.
%
Recently, many works use NAS to design GNN models due to the complexity of GNN~\cite{wang2022automated}. 
PolicyGNN~\cite{lai2020policy} uses reinforcement learning to train meta-strategies and then adaptively determines the choice of aggregation layers for each node.
SANE~\cite{zhao2020simplifying} and SNAG~\cite{zhao2020simplifying} search for aggregation functions using microscope-based and reinforcement learning-based strategies, respectively. 
The architecture-level approaches such as GraphNAS~\cite{gao2019graphnas}, AutoGNN~\cite{zhou2019auto}, and PSP~\cite{zhu2022psp} aim to search for architectural representations of each layer, including sampling functions, attention computation functions, aggregation functions, and activation functions.
The above works are based on homogeneous graphs.
Due to the rich semantic and structural information in heterogeneous graphs, applying NAS to heterogeneous graphs is more challenging. 
Recently, there exist some excellent attempts.
GEMS~\cite{han2020genetic} uses the evolutionary algorithm to search for meta-graphs between source and target nodes.
DiffMG~\cite{ding2021diffmg} uses differentiable methods to find the best meta-structures in heterogeneous graphs. 
However, the above works only focus on the GNN model and ignore the heterogeneous graph data itself, which is even more important in practice.

\subsection{Proximal Iteration}
Proximal iteration~\cite{parikh2014proximal} is used to handle the optimization problem with a constraint $\mathcal{C}$, i.e., $\mathop{\min}_x  f(x), \text{s.t. } x \in \mathcal{C}$, 
where $f$ is a differentiable objective function.
The proximal step is:
 \begin{align}
 \begin{split}
     &x^{(k+1)} = \text{prox}_{\mathcal{C}} \left( x^{(k)} - \epsilon \nabla f \left(x^{(k)} \right)\right) \\
      \ &\text{prox}_{\mathcal{C}}(x)  = \mathop{\arg \min}_z \frac{1}{2} (\lVert z - x \rVert)^2, \text{s.t. } z \in \mathcal{C}   
 \end{split}
 \end{align}
where $\epsilon$ is the learning rate. 
%
Due to the excellent theoretical guarantee and good empirical performance, proximal iteration has been applied to many deep learning problems (e.g., architecture search~\cite{yao2020efficient}).

\section{Preliminaries}
\label{preliminary}

\noindent\textbf{\emph{Heterogeneous Graph.}}  Given a graph $G = \langle V,E \rangle$ where $V$ and $E$ denote the node set and the edge set respectively, $G$ is heterogeneous when the number of node and edge types
exceeds 2. 
Each node $v \in V$ and each edge $e \in E$ are associated with a node type and an edge type respectively.

\vspace{2ex}
\noindent\textbf{\emph{Attribute Missing in Heterogeneous Graph.}} Let $x_v \in \mathbb{R}^d$ denote the original $d$-dimensional attribute vector in the node $v$. 
In practice, the attributes of some types of nodes are not available. 
Thus, the node set $V$ in $G$ can be divided into two subsets, i.e., $V^+$ and $V^-$, which denote the attributed node-set and no-attribute node-set.

\vspace{2ex}
\noindent\textbf{\emph{Attribute Completion.}} 
Let $X=\{x_v \ |\  v \in V^+\}$ denote the input attribute set. 
Attribute completion aims to complete the attribute for each no-attribute node $v \in V^-$ by leveraging the available attribute information $X$ and the topological structure of $G$.
Let $x_v^{C}$ denote the completed attribute. 
Thus, after completion, the node attributes for the training of heterogeneous GNN is $X^{new} = X \cup X^C = \{x_v \ |\  v \in V^+\} \cup \{x_v^c \ |\  v \in V^-\}$.
In this paper, we aim to search for the optimal completion operation for each no-attribute node to improve the prediction performance of GNN models.
\section{The Proposed Methodology}

In this section,  We first present the proposed completion operation search space and then introduce the differentiable search strategy.
Moreover, we introduce the optimization techniques including discrete constraints and the auxiliary unsupervised graph node clustering task for further improving the search efficiency.

\subsection{Search Space of Attribute Completion Operation}
\label{s:space}

Due to the semantic differences between nodes, using a single attribute completion operation for all no-attribute nodes belonging to the same node type is not reasonable. 
The available completion operations should be diverse and we can select the most suitable completion operation for each node with missing attributes.
Thus, to capture both the node semantics and the topological structure information during the attribute completion process, we first propose an expressive completion operation search space, which consists of topology-dependent and topology-independent operations.

Specifically, the topology-dependent operations employ the topology information of the graph to guide the attribute completion. 
Inspired by the node aggregation operations in typical GNNs (e.g., GraphSage~\cite{hamilton2017inductive}, GCN~\cite{kipf2017gcn}, APPNP~\cite{klicpera2018predict}), we design three topology-dependent attribute completion operations, i.e., mean, GCN-based, PPNP-based operations.
In contrast, the topology-independent operation directly uses one-hot encoding to replace the missing attribute.
\autoac{} aims to search the optimal operation for each no-attribute node from the general and scalable search space where we can draw on more node aggregation operations in GNNs as attribute completion operations.

\subsubsection{Topology-Dependent Completion Operation}
Such type of completion operations can be further divided into two categories: local attribute aggregation and global (i.e., multi-hop) attribute aggregation.

\noindent\textbf{\emph{Local Attribute Aggregation.}} Similar to the node aggregation in GraphSage~\cite{hamilton2017inductive}, we first propose mean attribute aggregation.

\emph{Mean Attribute Aggregation.}
For the node $v \in V^-$, we calculate the mean of neighbors' attributes to complete the missing attribute.
The completed attribute $x_v^{C}$ is as follows:
\begin{equation}
x_{v}^{C}=W \cdot \operatorname{mean}\left\{x_{u}, \forall u \in N_v^{+}\right\}
\end{equation}
where $N_v^{+}$ denotes the local (i.e, 1-hop) neighbors of node $v$ in set $V^+$. $W$ is the trainable transformation matrix.

\emph{GCN-based Attribute Aggregation.} Similar to spectral graph convolutions in GCN~\cite{kipf2017gcn}, we complete the missing attribute with the following renormalized graph convolution form.

\begin{equation}
x_{v}^{C}=\sum_{u \in N_v^{+}}(\operatorname{deg}(v) \cdot \operatorname{deg}(u))^{-1 / 2} \cdot x_{u} \cdot W
\end{equation}

\noindent\textbf{\emph{Global Attribute Aggregation.}}
Motivated by the node aggregation in APPNP~\cite{klicpera2018predict}, we propose PPNP-based completion operation for global attribute aggregation.

\emph{PPNP-based Attribute Aggregation.} Besides the GCN-based attribute completion, we use another popular node aggregation method PPNP (i.e., Personalized PageRank~\cite{klicpera2018predict}) for attribute completion. 
Specifically, let $A \in \mathbb{R}^{n \times n}$ denote the adjacency matrix of the graph $G$. $\tilde{A} = A+I_n$ denotes the adjacency matrix with added self-loops.
The form of PPNP-based attribute completion is:
\begin{equation}
 \begin{aligned}
  & X^{ppnp}=\alpha\left(I_{n}-(1-\alpha \hat{\tilde{A}})\right)^{-1} \cdot X^{\prime}, X^{\prime}=X \cdot W \\
  & X_{C}=\{X^{ppnp}_i \ |\ \forall i \in V^-\}
 \end{aligned}
\end{equation}
where $\hat{\tilde{A}} = \tilde{D}^{-1/2}\tilde{A}\tilde{D}^{-1/2}$ is the symmetrically normalized
adjacency matrix with self-loops, with the diagonal degree matrix $\tilde{D}$. 
$\alpha \in (0,1]$ is the restart probability.
Note that the missing attributes are filled with zeros in $X$.
After PPNP-based attribute aggregation, we complete the attributes of the nodes in $V^-$ with $X^{ppnp}$.

\subsubsection{Topology-Independent Completion Operation}
For the no-attribute nodes that have few neighbors or are less affected by the neighbor information, we can directly use one-hot encoding to replace the missing attributes. 
The one-hot representation of a specific node type is also a commonly used handcrafted attribute completion method~\cite{lv2021we}.
%
\textcolor{ black}{For example, there are $K$ distinct actors in IMDB.
The one-hot representation for the actor node is a $K$-dimensional vector.
For a specific actor, the element in the corresponding index is 1 and the others are 0.
Then, the one-hot representation is transformed linearly for dimension alignment. 
}

\subsubsection{Search Space Size Analysis}
In summary, the proposed search space $\mathcal{O}$ contains a diverse set of attribute completion operations.
Let $N^-$ denote the total number of nodes with missing attributes.
Thus, the space size can be calculated by ${\left| \mathcal{O} \right|}^{N^-}$, which is exponential to ${N^-}$.
In practice, the attribute missing of some node types is a common problem, leading to huge search space.
Thus, the block-box optimization-based search method (e.g., evolutionary algorithm) over a discrete search space is infeasible.
To address this issue, we propose a differentiable search strategy to find the optimal completion operations efficiently.

\subsection{Differentiable Search Strategy}
In this section, we first introduce a continuous relaxation scheme for the completion operation search space to make the search process to be differentiable.
Then, we introduce the differentiable search algorithm and two optimization techniques to improve the search efficiency.

\subsubsection{Continuous Relaxation and Optimization}
Inspired by the success of the differentiable NAS, we first design a continuous search space and then perform differentiable completion operation search via gradient descent. 

As shown in Equation~\ref{eq_nas_op_sum}, instead of searching over the discrete space, we view the completion operation as a weighted mixture of candidate choices.

\begin{equation}
    x_v^{C}= \sum_{o \in \mathcal{O}}{\frac{\exp \left(\alpha_{o}^{(v)}\right)}{\sum_{o^{\prime} \in \mathcal{O}} \exp \left(\alpha_{o^{\prime}}^{(v)}\right)} o\left( v\right)}
    \label{eq_nas_op_sum}
\end{equation}

where $v$ denotes the node with the missing attribute, $o$ denotes the candidate operation in the search space $\mathcal{O}$, $o\left( v\right)$ denotes the completed attribute of node $v$ with $o$. 
$\alpha^{\left( v \right)}$ indicates the mixing weight vector of dimension $\left| \mathcal{O} \right|$ for node $v$.
Furthermore, we refer to $\alpha = \{\alpha^{\left( v \right)} \ |\ v\in V^-\} \in \mathbb{R}^{N^- \times \left| \mathcal{O} \right|}$ as the completion parameters.

After continuous relaxation, the search objective becomes the learning of the completion parameters $\alpha$.
To this end, we formulate the search problem as an optimization problem that can jointly learn the completion
parameters $\alpha$ and the weights $w$ in the heterogeneous GNN by gradient descent.
Let $\mathcal{L}_{train}$ and $\mathcal{L}_{val}$ denote the training loss and validation loss respectively.
Since both losses are determined by the completion
parameters $\alpha$ and the weights $w$, the search objective is a bi-level optimization problem.

\begin{equation}
    \begin{gathered}
        \min _{\alpha} \mathcal{L}_{val}\left(\omega^{*}, \alpha\right) \\
        \text { s.t. }\omega^{*}=\operatorname{argmin}_w \mathcal{L}_{train}(\omega, \alpha)
    \end{gathered}
    \label{eq_nas_def} 
\end{equation}
where the upper-level optimization is for the optimal completion parameters $\alpha$ and the lower-level optimization is for the optimal weights $w$ in the GNN model.

\begin{figure*}
	\centering
	\includegraphics[width=1\textwidth]{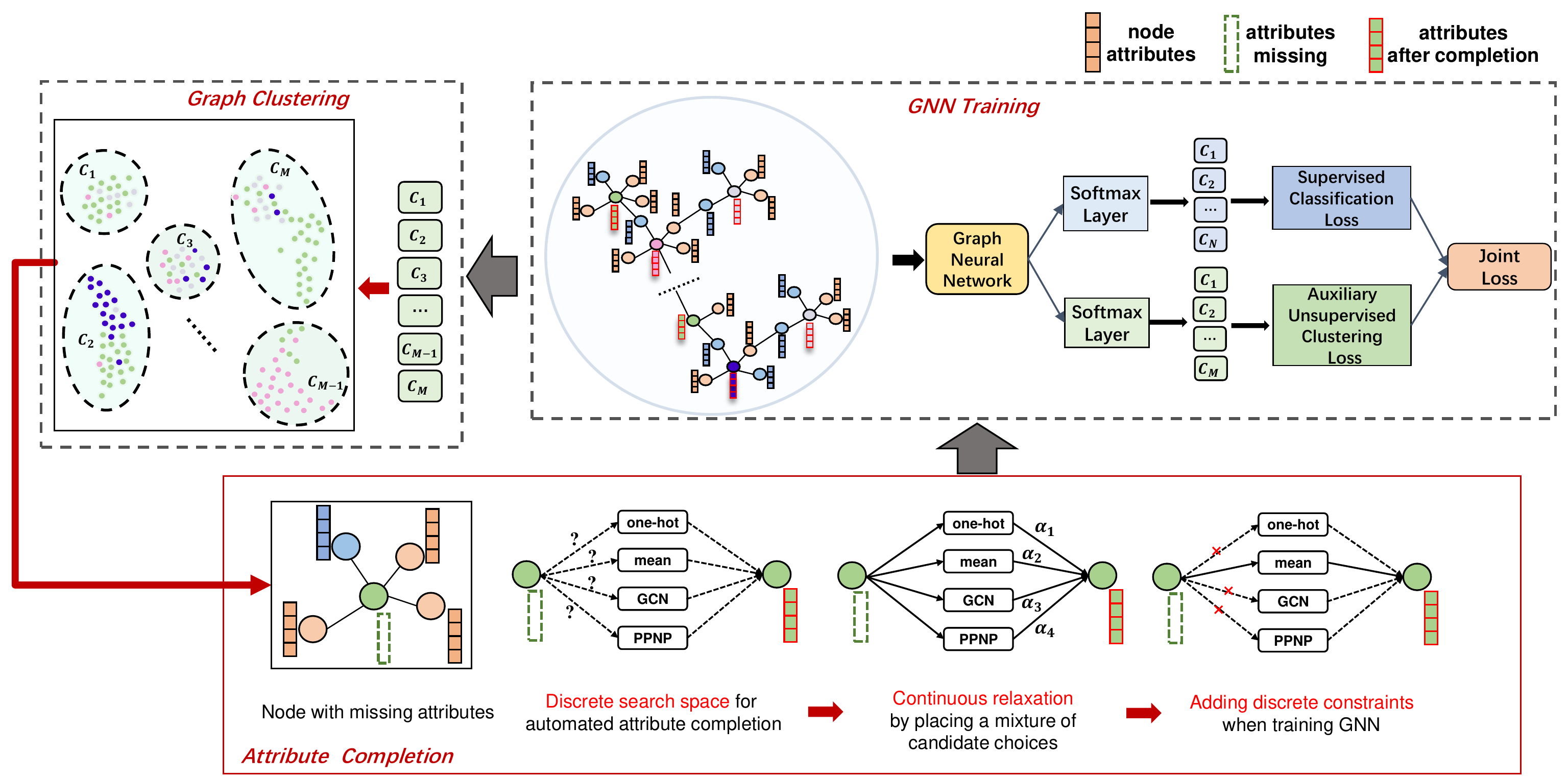}
		\vspace{-4ex}
	\caption
	{
		The overall workflow of automated attribute completion for the heterogeneous graph neural network.
	}
	\label{fig:framework}
	\vspace{-3ex}
\end{figure*}

\subsubsection{Overview}
Figure~\ref{fig:framework} shows the overall framework of automated attribute completion for heterogeneous graphs.
First, we perform a continuous relaxation of the search space by placing a mixture of candidate completion operations.
Then, the completion parameters $\alpha$ are optimized.
After determining the attribute completion operations for each no-attribute node, we view the completed attributes together with the raw attributes as the initial embedding for the training of the graph neural network.

\noindent\textbf{\emph{Why not use the weighted mixture.}}
Although the continuous relaxation allows the search of completion operations to be differentiable, there still exist following limitations when directly using the weighted mixture of all completion operations:

\begin{enumerate}
    \item \textit{High computational overhead:} After continuous relaxation, we need to perform all candidate completion operations for each no-attribute node when training heterogeneous GNNs, leading to huge computational overhead. Also, solving the bi-level optimization problem in Equation~\ref{eq_nas_def} incurs significant computational overhead.
    \item \textit{Performance gap:} At the end of the search,  continuous parameters $\alpha$ needs to be discretized, i.e., $\operatorname{argmax}_{o \in \mathcal{O}}{\alpha_{o}^{\left( v \right)}}$, resulting in inconsistent performance between searched and final completion operations.
    \item \textit{Large dimension of $\alpha$:} The dimension of completion parameters $\alpha$ is $N^- \times \left| \mathcal{O} \right|$, which is proportional to the total number of nodes with missing attributes. The large dimension of $\alpha$ leads to a slow convergence rate and low search efficiency.
\end{enumerate}

To address the first two issues (i.e., reducing computational overhead and avoiding performance gap), we first propose an efficient search algorithm with discrete constraints.
Specifically, for each no-attribute node $v$, the completion parameters satisfy the following constraints: $\alpha^{\left( v \right)} \in \mathcal{C} = \mathcal{C}_1 \cap \mathcal{C}_2$, where $\mathcal{C}_1 = \{ \alpha^{\left( v \right)} \mid \| \alpha^{\left( v \right)} \|_{0}=1 \}$, $\mathcal{C}_2 = \{ \alpha^{\left( v \right)} \mid  0 \leq \alpha^{\left( v \right)}_{i} \leq 1\}$.
The constraint $\mathcal{C}_2$ allows $\alpha$ to be optimized continuously, and $\mathcal{C}_1$ keeps the choices of completion operation to be discrete when training GNN.
As shown in Figure~\ref{fig:framework}, there is only one activated edge for each choice when training GNN, removing the need to perform all candidate completion operations. 
%
%
The final completion operation is derived from the learned completion parameter $\alpha$. For node $v$, the edge with the
maximum completion parameter will be kept. 
We leverage proximal iteration~\cite{parikh2014proximal} to solve the constrained optimization problem. 
Moreover, proximal iteration can improve the computational efficiency of optimizing $\alpha$ without second-order derivative.

Moreover, to address the third issue (i.e., reducing the dimension of $\alpha$), we propose an auxiliary unsupervised clustering task.
In practice, the no-attribute nodes with similar semantic characteristics may have the same completion operation.
Take the actor nodes in the IMDB dataset as an example. 
For the actors with a large number of representative movies, the average attribute aggregation operation is more suitable.
Thus, we can cluster all no-attribute nodes into $M$ clusters, where the nodes in each cluster have the same completion operation.
The optimization goal becomes to search for the optimal attribute completion operation for each cluster. 
In this way, the size of the completion parameters $\alpha$ is reduced from $N^{-} \times \left|\mathcal{O}\right|$ to $M \times \left|\mathcal{O}\right|$, $M \ll N^-$.
As shown in Figure~\ref{fig:framework}, the auxiliary unsupervised clustering loss can be jointly optimized with the node classification loss (i.e., cross-entropy).

The proposed framework \autoac{} is composed of multiple iterations.
In each iteration, the completion parameters $\alpha$ and the weights in the GNN are optimized alternatively.
Next, we introduce the search algorithm with discrete constraints and the auxiliary unsupervised clustering task in detail.

\subsection{Search Algorithm with Discrete Constraints}
Equation~\ref{eq_nas_def} implies a bi-level optimization problem with $\alpha$ as the upper-level variable and $w$ as the lower-level variable. 
Following the commonly used methods in meta learning~\cite{finn2017maml} and NAS~\cite{liu2018darts}, we use a one-step gradient approximation to the optimal internal weight parameters $\omega^*$ to improve the efficiency. 
Thus, the gradient of the completion parameters $\alpha$ is as follows (we omit the step index $k$ for brevity):
\begin{equation}
    \begin{aligned}
        & \nabla_{\alpha} \mathcal{L}_{val}\left(\omega^{*}, \alpha\right) \\
        \approx & \nabla_{\alpha} \mathcal{L}_{val}\left(\omega-\xi \nabla_{\omega} \mathcal{L}_{train}(\omega, \alpha), \alpha\right) \\
         =& \nabla_{\alpha} \mathcal{L}_{v a l}\left(\omega^{\prime}, \alpha\right)-\xi \nabla_{\alpha, \omega}^{2} \mathcal{L}_{train}(\omega, \alpha) \nabla_{\omega^{\prime}} \mathcal{L}_{val}\left(\omega^{\prime}, \alpha\right)
    \label{eq_one_step}
    \end{aligned}
\end{equation}
where $\omega$ is the weights of the GNN, $\xi$ is the learning rate of internal optimization, and $\omega^{\prime}=\omega-\xi \nabla_{\omega} \mathcal{L}_{train}(\omega, \alpha)$ indicates the weights for a one-step forward model.
we update the completion parameters $\alpha$ to minimize the validation loss.
In Equation~\ref{eq_one_step}, there exists a second-order derivative, which is expensive to compute due to a large number of parameters.
Also, the continuous relaxation trick further leads to huge computational overhead since all candidate completion operations need to be performed when training the GNN.
Moreover, the overall search process is divided into two stages: search and evaluation. 
In the evaluation stage, the continuous completion parameters $\alpha$ need to be discretized for replacing every mixed choice as the most likely operation by taking the argmax, leading to performance gap between the search and evaluation stage.

To optimize $\alpha$ efficiently and avoid the performance gap, we propose a search algorithm with discrete constraints when optimizing completion parameters $\alpha$. 
For the no-attribute node $v$, let the feasible space of $\alpha^{\left( v \right)}$ be $\mathcal{C} = \{ \alpha^{\left( v \right)} \mid \| \alpha^{\left( v \right)} \|_{0}=1 \wedge 0 \leq \alpha^{\left( v \right)}_{i} \leq 1\}$.
We denote it as the intersection of two feasible spaces (i.e., $\mathcal{C} = \mathcal{C}_1 \cap \mathcal{C}_2)$, where $\mathcal{C}_1 = \{ \alpha^{\left( v \right)} \mid \| \alpha^{\left( v \right)} \|_{0}=1 \}$, $\mathcal{C}_2 = \{ \alpha^{\left( v \right)} \mid  0 \leq \alpha^{\left( v \right)}_{i} \leq 1\}$.
The optimization problem under constraints can be solved by the proximal iterative algorithm.

\vspace{1ex}
\begin{prop}\label{prop:proximal}
$\text{prox}_{\mathcal{C}} (z) = \text{prox}_{\mathcal{C}_2} (\text{prox}_{\mathcal{C}_1} (z))$
\vspace{1ex}
\end{prop}
Inspired by Proposition~\ref{prop:proximal} ~\cite{parikh2014proximal,yao2020efficient}, in the $k$-th proximal iteration, we first get discrete variables constrained by $\mathcal{C}_1$, i.e., ${\bar{\alpha}}^{(k)} = \text{prox}_{\mathcal{C}_1}( {\alpha}^{(k)})$ (the node notation 
$v$ is omitted for brevity).
Then, we derive gradients w.r.t ${\bar{\alpha}}^{(k)}$ and keep $\alpha$ to be optimized as continuous variables but constrained by $\mathcal{C}_2$. 

\begin{equation}
    \alpha^{(k+1)}=\operatorname{prox}_{c_2}(\alpha^{(k)}-\epsilon \nabla_{\bar{\alpha}^{(k)}} \mathcal{L}_{v a l}(\bar{\alpha}^{(k)}))
    \label{eq_pa_iter}
\end{equation}

\begin{algorithm}[t]
\caption{Search Algorithm in \autoac{}}
\label{alg}
\begin{algorithmic}[1]
\STATE Initialize completion parameters $\alpha$ according to defined search space $\mathcal{O}$;
    \WHILE{ \textit{not converge} }
        \STATE Get $\textit{discrete}$ choices of attribute completion operations:
        $\bar{\alpha}^{(k)}=\operatorname{prox}_{c_1}(\alpha^{(k)})$
        
        \STATE Update $\alpha$ for \textit{continuous} variables: $\alpha^{(k+1)}=\operatorname{prox}_{c_2}(\alpha^{(k)}-\epsilon \nabla_{\bar\alpha^{(k)}} \mathcal{L}_{v a l}(\omega^{(k)}, \bar\alpha^{(k)}))$
        
        \STATE Refine \textit{discrete} choices after updating: $\bar{\alpha}^{(k+1)}=\operatorname{prox}_{c_1}(\alpha^{(k+1)})$
        
        \STATE Update $\omega^{(k)}$ by $\nabla_{\omega^{(k)}} \mathcal{L}_{train}(\mathbf{\omega}^{(k)},\bar{\alpha}^{(k+1)})$
        
    \ENDWHILE
\end{algorithmic}
\end{algorithm}

The detailed search algorithm is described in Algorithm~\ref{alg}.
First, we get a discrete representation of $\alpha$ by proximal step (Line 3).
Then, we view $\omega^{(k)}$ as constants and optimize $\alpha^{(k+1)}$ for continuous variables (Line 4). 
Since there is no need to compute the second-order derivative, the efficiency of updating $\alpha$ can be improved significantly.
After updating $\alpha$, we further refine discrete choices and get $\bar\alpha^{(k+1)}$ for updating $\omega^{(k)}$ on the training dataset, which contributes to reducing the performance gap caused by discretizing completion parameters $\alpha$ from continuous variables.
Moreover, since only one candidate choice is activated for each no-attribute node, the computational overhead can also be reduced.
%
The computational efficiency of updating $\alpha$ can be significantly improved.

\subsection{Auxiliary Unsupervised Clustering Task}
As mentioned before, the dimension of the completion parameters $\alpha$ is $N^- \times \left| \mathcal{O} \right|\left( \left|\mathcal{O}\right| \ll N^-, \left|\mathcal{O}\right| = 4 \right)$.
Take the DBLP dataset as an example, the number of nodes with missing attributes is about $1.2 \times {10}^4$, leading to a large dimension of completion parameters $\alpha$.
As a result, optimizing $\alpha$ with a limited size of validation dataset is very difficult.  

Inspired by the observation that the no-attribute nodes with similar explicit topological structure or implicit semantic characteristics, we further propose an auxiliary unsupervised clustering task to divide all no-attribute nodes into $M$ clusters.
In each cluster, all nodes share the same completion operation.
In this way, the dimension of the completion
parameters $\alpha$ can be reduced to $M \times \left|\mathcal{O}\right|$, $M \ll N^-$, and optimizing $\alpha$ becomes feasible and efficient.

It is well known that the EM algorithm~\cite{dempster1977maximum} is a commonly used method (e.g., K-Means\cite{macqueen1967classification}) to solve the problem of unsupervised clustering.
In the scenario of graph node clustering, let $h_v$ denote the hidden node representation learned by the heterogeneous GNN.
The E-step is responsible for assigning the optimal cluster for each node $v$ by calculating the distances between $h_v$ and all cluster centers.
The M-step is used to update the centers of all clusters.
The E-step and M-step are performed alternately until convergence.

Although the EM algorithm has a convergence guarantee, it is sensitive to the initial values, making it difficult to apply to the proposed automated completion framework.
The main reason is that the bi-level optimization problem defined in Equation~\ref{eq_nas_def} is iterative.
In the early optimization process, the weights of the GNN have not yet converged and the node representations learned in the GNN are less informative.
Such low-quality representations lead to inaccurate clustering, which has a negative impact on the subsequent clustering quality and further leads to a deviation from the overall optimization direction.

To address this issue, we first formulate the problem of unsupervised node clustering as a form of soft classification, and use the assignment matrix $\boldsymbol{C}$ to record the probability of each node belonging to each cluster. 
Moreover, as shown in Figure~\ref{fig:framework}, we embed the clustering process into the bi-level iterative optimization process.

Motivated by graph pooling and graph module partitioning, we introduce the Spectral Modularity Function $\mathcal{Q}$~\cite{good2010performance}\cite{bianchi2020spectral}. From a statistical perspective, this function can reflect the clustering quality of graph node modules through the assignment matrix $\boldsymbol{C}$~\cite{tsitsulin2020graph}:

\begin{equation}
    \mathcal{Q}=\frac{1}{2 \left| E \right|} \sum_{i j}\left[\boldsymbol{A}_{i j}-\frac{d_{i} d_{j}}{2 \left| E \right|}\right] \delta\left(c_{i}, c_{j}\right)
    \label{eq_spectral_q}
\end{equation}
where $\left| E \right|$ is the number of edges in the graph, $\delta(c_i, c_j)=1$ only if nodes $i$ and $j$ are in the same cluster, otherwise 0.
$d_i$ and $d_j$ represent the degrees of node $i$ and node $j$ respectively. It can be known that in a random graph, the probability that node $i$ and node $j$ are connected is $\frac{d_{i} d_{j}}{2 \left| E \right|}$~\cite{tsitsulin2020graph}.

Then, the optimization goal is converted into maximizing the spectral modularity function $\mathcal{Q}$, but it is an NP-hard problem. 
Fortunately, this function can be represented by an approximate spectral domain relaxation form:

\begin{equation}
    \mathcal{Q}=\frac{1}{2 \left| E \right|} \operatorname{Tr}\left(\boldsymbol{C}^{\top} \boldsymbol{B} \boldsymbol{C}\right)
    \label{eq_spectral_q_convert}
\end{equation}
where $\mathcal{C}_{i j} \in[0,1]$ denotes the cluster probability. $\boldsymbol{B}$ is the modular matrix $\boldsymbol{B}=\boldsymbol{A}-\frac{\boldsymbol{d} \boldsymbol{d}^{\top}}{2 \left| E \right|}$. 
Finding the optimal solution of the assignment matrix $\boldsymbol{C}$ is to maximize $\mathcal{Q}$.
%
To prevent falling into local optimum (i.e., all nodes tend to be in the same cluster), we further add the collapse regularization term.
The assignment matrix $\boldsymbol{C}$ should be amortized as adaptively as possible, so as to skip the local optimum.

Let $\mathcal{L}_{G m o C}$ denote the unsupervised clustering loss, which can be expressed as:
\begin{equation}
    \mathcal{L}_{G m o C}=\underbrace{-\frac{1}{2 \left| E \right|} \operatorname{Tr}\left(\boldsymbol{C}^{\top} \boldsymbol{B C}\right)}_{\text{modularity loss}}+\underbrace{\frac{\sqrt{M}}{\left| V \right|}\left\|\sum_{i} \boldsymbol{C}_{i}^{\top}\right\|_{F}}_{\text{collapse regularization}}
    \label{eq_gmoc_loss}
\end{equation}
where $\left| V \right|$ is the number of nodes, $M$ is the number of clusters, $\|\cdot\|_{F}$ represents the Frobenius norm of the matrix.
Note that $\mathcal{L}_{G m o C}$ can be jointly optimized with the supervised classification loss.
Specifically, $\mathcal{L}_{G m o C}$ can be used as an auxiliary task for the bi-level optimization problem in Equation~\ref{eq_nas_def}. 
The unsupervised clustering loss is added to $\mathcal{L}_{train}$ for joint optimization. 
Let $\lambda$ denote the loss-weighted coefficient. 
The optimization objective is updated as:

\begin{equation}
    \begin{aligned}
        & \min _{\alpha} \mathcal{L}_{v a l}\left(w^{*}, \alpha\right) \\
        & \text { s.t. } w^{*}=\arg \min _{w}\left(\mathcal{L}_{train}(w, \alpha)+\lambda \mathcal{L}_{G m o C}\right) \\
    \end{aligned}
    \label{eq_total_loss}
\end{equation}

\subsection{\textcolor{ black}{Complexity Analysis}}
In the heterogeneous graph $G = \langle V, E \rangle$, the total number of nodes is $N$, the total number of nodes with missing attributes is $N^-$, and the embedding dimension is $k$.
In each iteration of Equation~\ref{eq_total_loss}, we can divide the search process of \autoac{} into three phases, i.e., attribute completion phase, upper-level optimization for completion parameters $\alpha$, and lower-level optimization for weights $\omega$.
%
%
%
%
We first analyze the computational complexity.
Since discrete constraints are performed, only one candidate completion operation is activated for each no-attribute node.
%
%
The computational complexity of each completion operation is as follows: Mean attribute aggregation: $O(N^- \times k^2)$, GCN-based attribute aggregation: $O(N^- \times k^2)$, PPNP-based attribute aggregation: $O(N \times k^2)$, one-hot attribute completion: $O(1)$. 
Thus, the computational complexity of the attribute completion phase is $O(N \times k^2)$.
In the upper-level optimization phase, the complexity is $O(C_{H}+\left|\mathcal{O}\right| \times M \times b_\alpha)$, where $C_{H}$ denotes the forward computation overhead of the heterogeneous GNN, $b_\alpha$ the gradient computation overhead for each completion parameter.
For brevity, we omit the difference between the validation and training datasets.
The lower-level optimization phase contains the optimization of weights and unsupervised clustering.
The complexity of optimizing $\omega$ is $O(C_{H}+\left| \omega \right| \times b_\omega)$, where $b_\omega$ is the gradient computation overhead for each weight parameter.
The complexity of calculating the clustering loss $\mathcal{L}_{G m o C}$ is $O(d^2 \times N + \left| E \right|)$~\cite{tsitsulin2020graph}, where $d$ is the average degree.
Overall, the computational complexity of each iteration 
is, $O(N \times k^2) + O(C_{H}+ \left|\mathcal{O}\right| \times M \times b_\alpha) + \left| \omega \right| \times b_\omega) + O(d^2 \times N + \left| E \right|)$.

Next, we analyze the space complexity of \autoac{}.
For the attribute completion phase, the space complexity is $O(k^2)$.
For the optimization phase, the space complexity is $O(N \times k + \left|\mathcal{O}\right| \times M + \left| \omega \right| + N \times M)$, where $O(N \times M)$ is the space complexity in the unsupervised clustering. 

%
%

\section{Experiments}


\subsection{Experimental Setup}

\begin{table}
  \caption{Statistics of the datasets}
  \label{Statistics of the datasets}
  \setlength\tabcolsep{3pt}
  \centering
  \scalebox{0.9}{
  \begin{tabular}{ccclccc}
    \toprule
    Datasets     & \#Nodes     & \makecell{\#Node \\ Types} & \makecell{\#Nodes under \\ Each Type} &\#Edges &\makecell{Target Node/Edge \\ Type} &Attribute \\
    \midrule
    \multirow{4}{*}{DBLP} & \multirow{4}{*}{26128}  & \multirow{4}{*}{4}  &author(A):4057 & \multirow{4}{*}{239566}  & \multirow{3}{*}{author} &A:Missing  \\
    \multirow{4}{*}{} & \multirow{4}{*}{}  & \multirow{4}{*}{}  &paper(P):14328 & \multirow{4}{*}{}  & \multirow{3}{*}{paper-author} &P:Raw  \\
    \multirow{4}{*}{} & \multirow{4}{*}{}  & \multirow{4}{*}{}  &term(T):7723 & \multirow{4}{*}{}  & \multirow{3}{*}{} &T:Missing  \\
    \multirow{4}{*}{} & \multirow{4}{*}{}  & \multirow{4}{*}{}  &venue(V):20 & \multirow{4}{*}{}  & \multirow{3}{*}{} &V:Missing  \\
    \midrule
    \multirow{4}{*}{ACM} & \multirow{4}{*}{10942}  & \multirow{4}{*}{4}  &paper(P):3025 & \multirow{4}{*}{547872}  & \multirow{4}{*}{paper} &P:Raw  \\
    \multirow{4}{*}{} & \multirow{4}{*}{}  & \multirow{4}{*}{}  &author(A):5959 & \multirow{4}{*}{}  & \multirow{4}{*}{} &A:Missing  \\
    \multirow{4}{*}{} & \multirow{4}{*}{}  & \multirow{4}{*}{}  &subject(S):56 & \multirow{4}{*}{}  & \multirow{4}{*}{} &S:Missing  \\
    \multirow{4}{*}{} & \multirow{4}{*}{}  & \multirow{4}{*}{}  &term(T):1902 & \multirow{4}{*}{}  & \multirow{4}{*}{} &T:Missing  \\
    \midrule
    \multirow{4}{*}{IMDB} & \multirow{4}{*}{21420}  & \multirow{4}{*}{4}  &movie(M):4932 & \multirow{4}{*}{86642}  & \multirow{3}{*}{movie} &M:Raw  \\
    \multirow{4}{*}{} & \multirow{4}{*}{}  & \multirow{4}{*}{}  &director(D):2393 & \multirow{4}{*}{}  & \multirow{3}{*}{movie-keyword} &D:Missing \\
    \multirow{4}{*}{} & \multirow{4}{*}{}  & \multirow{4}{*}{}  &actor(A):6124 & \multirow{4}{*}{}  & \multirow{3}{*}{} &A:Missing  \\
    \multirow{4}{*}{} & \multirow{4}{*}{}  & \multirow{4}{*}{}  &keyword(K):7971 & \multirow{4}{*}{}  & \multirow{3}{*}{} &K:Missing  \\
    \midrule
    \multirow{3}{*}{LastFM} & \multirow{3}{*}{20612}  & \multirow{3}{*}{3}  &user(U):1892 & \multirow{3}{*}{141521}  & \multirow{3}{*}{user-artist} &U:Missing  \\
    \multirow{3}{*}{} & \multirow{3}{*}{}  & \multirow{3}{*}{}  &artist(A):17632 & \multirow{3}{*}{}  & \multirow{3}{*}{} &A:Raw \\
    \multirow{3}{*}{} & \multirow{3}{*}{}  & \multirow{3}{*}{}  &tag(T):2980 & \multirow{3}{*}{}  & \multirow{3}{*}{} &T:Missing  \\
    \bottomrule
  \end{tabular}}
  \vspace{-3ex}
\end{table}

\subsubsection{Experimental Setting}
We use the recently proposed Heterogeneous Graph Benchmark (HGB)~\cite{lv2021we} to conduct all experiments, which offers a fair way to compare heterogeneous GNN models. 
HGB gives a set of standard benchmark datasets and unified strategies for feature preprocessing and data split.
%
%
In the node classification task, all edges are available during training, and node labels are split according to 24\% for training, 6\% for validation, and 70\% for test in each dataset.
In the link prediction task, we mask 10\% edges of the target link type and the negative edges are randomly sampled.
%
The statistics of the four datasets are summarized in Table~\ref{Statistics of the datasets}.
More details of datasets can be seen in Appendix~\ref{ss:details}.

Moreover, the handcrafted attribute completion methods for existing heterogeneous GNNs are provided by HGB.
Micro-F1 and Macro-F1 are provided to evaluate the node classification performance, while the MRR and ROC-AUC metrics are used for link prediction.
The evaluation metrics are obtained by submitting predictions to the HGB website\footnote{https://www.biendata.xyz/competition/hgb-1/}.

\subsection{Implementation Details}
All experiments are performed in the transductive setting.
We employ the Adam optimizer~\cite{kingma2014adam} to optimize both $\omega$ and $\alpha$.
For optimizing $\omega$, the learning rate and the weight decay are 5e-4 and 1e-4 respectively.
$
$
For optimizing $\alpha$, the learning rate and the weight decay are 5e-3 and 1e-5 respectively.

We implement \autoac{} based on the widely-used heterogeneous GNNs, i.e., MAGNN~\cite{fu2020magnn} and SimpleHGN~\cite{lv2021we}.
The loss weighted coefficient $\lambda$ and the number of clusters $M$ are two hyperparameters of \autoac{}.
For MAGNN, we empirically set $\lambda$ to 0.5 for all datasets, $M$ to 4 for the DBLP and ACM datasets, 16 for the IMDB dataset.
For SimpleHGN, $\lambda$ is 0.4 for all datasets, and $M$ is 8 for the DBLP dataset, 12 for the ACM and IMDB datasets.
Moreover, all the GNN models are implemented with PyTorch. 
All experiments are run on a single GPU (NVIDIA Tesla V100) five times and the average performance and standard deviation are reported.

\subsection{Effectiveness of \autoac{}}

\begin{table*}[h]
  \caption{Performance and runtime (clock time in seconds) comparison between \autoac{} and SOTA humancrafted heterogeneous GNNs on node classification. The bold and the underline indicate the best and the second best in each category (i.e., using and not using meta-path). * indicates the global best in all models. $p$-value indicates the statistically significant improvement (i.e., t-test with $p < 0.05$) over the best baseline.}
  \centering
  \scalebox{0.65}{
  \label{tbl:compare_gnn}
  \begin{tabular}{l|cccc|cccc|cccc}
    \toprule
    Dataset   & \multicolumn{4}{c}{DBLP}  & \multicolumn{4}{c}{ACM}  & \multicolumn{4}{c}{IMDB} \\
    \midrule
   \multirow{2}{*}{}& Macro-F1 & Micro-F1 &\makecell{\textcolor{ black}{Runtime} 
 \\ \textcolor{ black}{(Total)}}&\makecell{\textcolor{ black}{Runtime} \\ \textcolor{ black}{(Per epoch)}}& Macro-F1 & Micro-F1 &\makecell{\textcolor{ black}{Runtime} 
 \\ \textcolor{ black}{(Total)}}&\makecell{\textcolor{ black}{Runtime} \\ \textcolor{ black}{(Per epoch)}}& Macro-F1 & Micro-F1&\makecell{\textcolor{ black}{Runtime} 
 \\ \textcolor{ black}{(Total)}}&\makecell{\textcolor{ black}{Runtime} \\ \textcolor{ black}{(Per epoch)}} \\
    \midrule
    HAN & \underline{93.17$\pm$0.19}	& 93.64$\pm$0.17	&\textcolor{ black}{44}&\textcolor{ black}{0.23}& 87.68$\pm$1.94	& 87.73$\pm$1.81 &	\textcolor{ black}{31}&\textcolor{ black}{0.25}&\textbf{59.70$\pm$0.90} &	\underline{65.61$\pm$0.54}& \textcolor{ black}{13}&\textcolor{ black}{0.08}\\
    GTN & 93.52$\pm$0.55	& 93.97$\pm$0.54	& \textcolor{ black}{13600}&\textcolor{ black}{340}& 91.63$\pm$1.27 &	91.53$\pm$1.30 & \textcolor{ black}{3234}&\textcolor{ black}{77}&	\underline{59.26$\pm$0.84}	& 64.07$\pm$0.65 & \textcolor{ black}{9960}&\textcolor{ black}{249}\\
     HetSANN& 84.08$\pm$1.01&	84.96$\pm$0.88& \textcolor{ black}{201}&\textcolor{ black}{0.93}&	90.09$\pm$1.06&	90.00$\pm$1.02& \textcolor{ black}{470}&\textcolor{ black}{1.50}&	49.25$\pm$0.57&	57.47$\pm$1.12 & \textcolor{ black}{520}&\textcolor{ black}{1.13}\\
     
      \textcolor{ black}{HGCA} & \textcolor{ black}{93.05$\pm$0.46}	&\textcolor{ black}{93.62$\pm$0.41}	&\textcolor{ black}{495}&\textcolor{ black}{55}& \underline{91.75$\pm$0.54}	& \underline{91.67$\pm$0.56} &	\textcolor{ black}{30}&\textcolor{ black}{1.5}&43.54$\pm$1.17 &	53.44$\pm$1.00& \textcolor{ black}{56}&\textcolor{ black}{2.8}\\
      
    MAGNN & 93.16$\pm$0.38 &	\underline{93.65$\pm$0.34} & \textcolor{ black}{401}&\textcolor{ black}{19}&	91.06$\pm$1.44 &	90.95$\pm$1.43 & \textcolor{ black}{230}&\textcolor{ black}{23}&	56.92$\pm$1.76 &	65.11$\pm$0.59 & \textcolor{ black}{108}&\textcolor{ black}{9.8}\\
    \textbf{MAGNN-\autoac{}} & \textbf{93.95$\pm$0.30} &	\textbf{94.39$\pm$0.25} & \textcolor{ black}{432}&\textcolor{ black}{21}&	\textbf{91.84$\pm$0.45} &	\textbf{91.77$\pm$0.45}& \textcolor{ black}{684}&\textcolor{ black}{25} &	58.96$\pm$1.31 &	\textbf{66.11$\pm$0.53}& \textcolor{ black}{576}&\textcolor{ black}{11}\\
    \midrule 
    HGT & 92.77$\pm$0.35	& 93.44$\pm$0.31 & \textcolor{ black}{131}&\textcolor{ black}{1.87}& 90.27$\pm$0.55 &	90.14$\pm$0.51 &  \textcolor{ black}{545}&\textcolor{ black}{7.07}&\underline{63.02$\pm$0.80} & 67.01$\pm$0.36 & \textcolor{ black}{257}&\textcolor{ black}{3.38}\\
    HetGNN &92.77$\pm$0.24&	93.23$\pm$0.23& \textcolor{ black}{20580}&\textcolor{ black}{98}&	84.93$\pm$0.78&	84.83$\pm$0.76& \textcolor{ black}{25410}&\textcolor{ black}{121}&	47.87$\pm$0.33&	50.83$\pm$0.26& \textcolor{ black}{18270}&\textcolor{ black}{87}\\
    GCN& 90.54$\pm$0.27&	91.18$\pm$0.25	& \textcolor{ black}{29}&\textcolor{ black}{0.09}&92.63$\pm$0.23&	92.60$\pm$0.22& \textcolor{ black}{26}&\textcolor{ black}{0.08}&	59.95$\pm$0.72&	65.35$\pm$0.35& \textcolor{ black}{10}&\textcolor{ black}{0.11} \\
    GAT &92.96$\pm$0.35&	93.46$\pm$0.35& \textcolor{ black}{14}&\textcolor{ black}{0.14}&	92.41$\pm$0.84&	92.39$\pm$0.84& \textcolor{ black}{29}&\textcolor{ black}{0.14}&	56.95$\pm$1.55&	64.24$\pm$0.55 & \textcolor{ black}{10}&\textcolor{ black}{0.21}\\
    SimpleHGN& \underline{93.83$\pm$0.18} &	\underline{94.25$\pm$0.19} & \textcolor{ black}{43}&\textcolor{ black}{0.39}&	\underline{92.92$\pm$0.67} &	\underline{92.85$\pm$0.68} & \textcolor{ black}{42}&\textcolor{ black}{0.47}&	62.98$\pm$1.66&	\underline{67.42$\pm$0.42} & \textcolor{ black}{25}&\textcolor{ black}{0.36}\\
    \textbf{SimpleHGN-\autoac{}}  & \textbf{95.15$\pm$0.29*}&	9\textbf{5.52$\pm$0.26*}& \textcolor{ black}{72}&\textcolor{ black}{0.58}&	\textbf{93.86$\pm$0.18*}&	\textbf{93.80$\pm$0.18*}& \textcolor{ black}{108}&\textcolor{ black}{0.62}&	\textbf{64.92$\pm$0.58*}&	\textbf{67.94$\pm$0.41*}& \textcolor{ black}{72}&\textcolor{ black}{0.55}\\
    \midrule
    $p$-value & \textbf{$2.9  \times e-8$} & \textbf{$ 3.3 \times e-9 $} &-&- &	\textbf{$1.6 \times e-6 $} & \textbf{$	2.9 \times e-6$} &-&-&	\textbf{$1.4 \times e-6$} &	\textbf{$9.8 \times e-6$} \\
    \bottomrule
  \end{tabular}}
\end{table*}

\begin{table*}[h]
  \caption{Performance comparison between \autoac{} and HGNNAC.The bold and the underlined indicate the best and the second best in each category. $p$-value indicates the statistically significant improvement (i.e., t-test with $p < 0.05$) over the best baseline.}
  \centering
  \scalebox{0.85}{
  \label{tbl:compare_hgnnac}
  \begin{tabular}{l|cc|cc|cc}
    \toprule
    Dataset   & \multicolumn{2}{c}{DBLP}  & \multicolumn{2}{c}{ACM}  & \multicolumn{2}{c}{IMDB} \\
    \midrule
    Model $\backslash$ Metrics & Macro-F1 & Micro-F1 & Macro-F1 & Micro-F1 & Macro-F1 & Micro-F1 \\
    \midrule
    MAGNN & \underline{93.16$\pm$0.38} &	\underline{93.65$\pm$0.34} &	\underline{91.06$\pm$1.44} &	\underline{90.95$\pm$1.43} &	\underline{56.92$\pm$1.76} &	\underline{65.11$\pm$0.59} \\
    MAGNN-HGNNAC&92.97$\pm$0.72	&93.43$\pm$0.69&	90.89$\pm$0.87&	90.83$\pm$0.87&	56.63$\pm$0.81&	63.85$\pm$0.85\\
    \textbf{MAGNN-AutoAC}& \textbf{93.95$\pm$0.30}&\textbf{94.39$\pm$0.25}&\textbf{91.84$\pm$0.45}&\textbf{91.77$\pm$0.45}&\textbf{58.96$\pm$1.31}&\textbf{66.11$\pm$0.53} \\
    \midrule
    SimpleHGN& \underline{93.83$\pm$0.18} &	\underline{94.25$\pm$0.19} &	92.92$\pm$0.67&	92.85$\pm$0.68&	62.98$\pm$1.66&	67.42$\pm$0.42\\
    SimpleHGN-HGNNAC & 93.24$\pm$0.49	& 93.73$\pm$0.45	& \underline{93.16$\pm$0.24} & \underline{93.09$\pm$0.23} &	\underline{64.44$\pm$1.13}	& \underline{67.67$\pm$0.39} \\
   \textbf{SimpleHGN-\autoac{}}  &\textbf{95.15$\pm$0.29}&	\textbf{95.52$\pm$0.26}&	\textbf{93.86$\pm$0.18}&	\textbf{93.80$\pm$0.18}&	\textbf{64.92$\pm$0.58}&	\textbf{67.94$\pm$0.41}\\
   \midrule
   $p$-value & $2.9 \times e ^{-8}	$& $3.3 \times e ^{-9}$	& $7.3 \times e ^{-7}$ &$1.1 \times e ^{-6} $&	$4 \times e ^{-3}$	& $1 \times e ^{-3}$ \\
    \bottomrule
  \end{tabular}}
\end{table*}
\begin{table*}[h]
  \caption{The overall runtime overhead (clock time in seconds) of \autoac{} and HGNN-AC. $/$ indicates that the stage is not involved.} 
  \centering
  \scalebox{0.85}{
  \label{tbl:time}
  
  \begin{tabular}{ccccccc}
    \toprule
    \multirow{2}{*}{Datasets}   & \multirow{2}{*}{Models}  &\multicolumn{4}{c}{End-to-End Runtime Overhead (Seconds)} &\multirow{2}{*}{Speedup}\\
     \multirow{2}{*}{}  & \multirow{2}{*}{}   & Pre-learn & Search &Train/Retrain & \textbf{Total} & \multirow{2}{*}{}  \\
    \midrule
    \multirow{4}{*}{DBLP} & SimpleHGN-HGNNAC  & 33048&/ &432  & 33480  & \multirow{2}{*}{\textbf{465$\times$}} \\
   \multirow{4}{*}{} & \textbf{SimpleHGN-\autoac{}}  &/ &	36&\textcolor{ black}{36}&	\textbf{\textcolor{ black}{72}} &\multirow{2}{*}{}  \\
   \cmidrule(r){2-7} 
   \multirow{4}{*}{} & MAGNN-HGNNAC  &33048&	/&	900&	33948  & \multirow{2}{*}{\textbf{78$\times$}} \\
   \multirow{4}{*}{} & \textbf{MAGNN-\autoac{}}  &/	&72&	\textcolor{ black}{360}& \textbf{\textcolor{ black}{432}}	 &\multirow{2}{*}{}\\
    \midrule
    \multirow{4}{*}{ACM} & SimpleHGN-HGNNAC  & 3888&/ &432  & 4320  & \multirow{2}{*}{\textbf{40$\times$}}  \\
   \multirow{4}{*}{} &\textbf{SimpleHGN-\autoac{}}  & /&	72&	\textcolor{ black}{36}&	\textbf{\textcolor{ black}{108}}  &\multirow{2}{*}{} \\
   \cmidrule(r){2-7} 
   \multirow{4}{*}{} & MAGNN-HGNNAC  &3888&	/&	1260&	5148   &\multirow{2}{*}{\textbf{7.5$\times$}}\\
   \multirow{4}{*}{} & \textbf{MAGNN-\autoac{}}  &/	&432&	\textcolor{ black}{252}&	\textbf{\textcolor{ black}{684}} &\multirow{2}{*}{}\\
    \midrule
     \multirow{4}{*}{IMDB} & SimpleHGN-HGNNAC  & 8568&/ &324  & 8892   &\multirow{2}{*}{\textbf{123$\times$}}\\
   \multirow{4}{*}{} & \textbf{SimpleHGN-\autoac{}}  & /&	36&	\textcolor{ black}{36}&	\textbf{\textcolor{ black}{72}}  &\multirow{2}{*}{} \\
   \cmidrule(r){2-7} 
   \multirow{4}{*}{} & MAGNN-HGNNAC  &8568&	/&	180&	8748   &\multirow{2}{*}{\textbf{15$\times$}}\\
   \multirow{4}{*}{} &\textbf{MAGNN-\autoac{}} &/	&504&	\textcolor{ black}{72}&\textbf{\textcolor{ black}{576}}	&\multirow{2}{*}{} \\
    \bottomrule
  \end{tabular}}
  \vspace{-2ex}
\end{table*}

\begin{table*}
  \caption{Performance and runtime (clock time in seconds) comparison on link prediction. The bold and the underline indicate the best and the second best. $p$-value indicates the statistically significant improvement (i.e., t-test with $p < 0.05$) over the best baseline.}
  \centering
   \scalebox{0.65}{
  \label{link prediction}
  \centering
  \begin{tabular}{l|cccc|cccc|cccc}
    \toprule
    Dataset   & \multicolumn{4}{c}{LastFM}  & \multicolumn{4}{c}{DBLP} & \multicolumn{4}{c}{IMDB} \\
    \midrule
   Model $\backslash$ Metrics & \textcolor{ black}{ROC-AUC}&\textcolor{ black}{MRR}&\makecell{\textcolor{ black}{Runtime} 
 \\ \textcolor{ black}{(Total)}}&\makecell{\textcolor{ black}{Runtime} \\ \textcolor{ black}{(Per epoch)}}&\textcolor{ black}{ROC-AUC}&\textcolor{ black}{MRR}&\makecell{\textcolor{ black}{Runtime} \\ \textcolor{ black}{(Total)}}&\makecell{\textcolor{ black}{Runtime} \\ \textcolor{ black}{(Per epoch)}}&\textcolor{ black}{ROC-AUC}&\textcolor{ black}{MRR}&\makecell{\textcolor{ black}{Runtime}
 \\ \textcolor{ black}{(Total)}}&\makecell{\textcolor{ black}{Runtime} \\ \textcolor{ black}{(Per epoch)}}\\
    \midrule
    GATNE & 66.87$\pm$0.16	& 85.93$\pm$0.63	& \textcolor{ black}{75960}&\textcolor{ black}{15435}&71.94$\pm$2.00	& 87.23$\pm$0.76&\textcolor{ black}{92160}&\textcolor{ black}{16278} & 47.45$\pm$6.48	& 74.58$\pm$3.34 &\textcolor{ black}{71280}&\textcolor{ black}{14269} \\
    HetGNN & 62.09$\pm$0.01	& 85.56$\pm$0.14&\textcolor{ black}{20580}&\textcolor{ black}{98}	& 88.89$\pm$0.40	& 94.39$\pm$0.62&\textcolor{ black}{22050}&\textcolor{ black}{105} & 56.55$\pm$0.83	& 78.10$\pm$0.56  &\textcolor{ black}{19950}&\textcolor{ black}{95}\\
    GCN & 59.17$\pm$0.31	& 79.38$\pm$0.65&\textcolor{ black}{13}&\textcolor{ black}{0.13}	& 80.48$\pm$0.81	& 90.99$\pm$0.56 &\textcolor{ black}{31}&\textcolor{ black}{0.12}& 51.90$\pm$1.10	& 76.99$\pm$1.87&\textcolor{ black}{28}&\textcolor{ black}{0.11} \\
    GAT & 58.56$\pm$0.66	& 77.04$\pm$2.11	&\textcolor{ black}{10}&\textcolor{ black}{0.12}& 72.89$\pm$3.09	& 82.56$\pm$3.35 &\textcolor{ black}{32}&\textcolor{ black}{0.15}& 48.30$\pm$1.35	& 76.74$\pm$2.00 &\textcolor{ black}{12}&\textcolor{ black}{0.10} \\
    SimpleHGN 	 &	 \underline{67.16$\pm$0.37} & 	\underline{86.73$\pm$0.27}&\textcolor{ black}{46}&\textcolor{ black}{0.35} &	\underline{94.61$\pm$0.11} &	
    \underline{97.21 $\pm$0.16} &\textcolor{ black}{58}&\textcolor{ black}{0.75}&	\underline{57.92$\pm$2.32} &	
    \underline{79.09 $\pm$1.40}&\textcolor{ black}{28}&\textcolor{ black}{0.44}\\
    \textbf{SimpleHGN-AutoAC} 	 &	 \textbf{67.72$\pm$0.17} & \textbf{87.10$\pm$0.19}&\textcolor{ black}{42}&\textcolor{ black}{0.43} &	\textbf{95.87$\pm$0.66} &	\textbf{98.21$\pm$0.21} &\textcolor{ black}{61}&\textcolor{ black}{0.87}&	\textbf{74.14$\pm$0.73} &	\textbf{86.27$\pm$0.45} &\textcolor{ black}{32}&\textcolor{ black}{0.49}\\
    \midrule
    $p$-value & $9.3 \times e ^{-4}	$& $9.5 \times e ^{-4}$&-&-	& $4.2 \times e ^{-5}$ &$8.2 \times e ^{-7} $&-&-&	$6.7 \times e ^{-9}$	& $7.2 \times e ^{-10}$ &-&-\\
    \bottomrule
  \end{tabular}}
  \vspace{-1ex}
\end{table*}

\subsubsection{Performance comparison with humancrafted heterogeneous GNNs}
Depending on whether or not the meta-path is used, we divide the humancrafted heterogeneous GNNs into two categories: 
\begin{itemize}
\item GNNs with meta-path: HAN~\cite{zhang2019heterogeneous}, GTN~\cite{yun2019graph},
HetSANN~\cite{hong2020attention}, MAGNN~\cite{fu2020magnn}, \textcolor{ black}{HGCA~\cite{9724614}}.
\item GNNs without meta-path: HGT~\cite{hu2020heterogeneous}, 
GATNE~\cite{cen2019representation}, HetGNN~\cite{zhang2019heterogeneous}, GCN~\cite{kipf2017gcn} and GAT~\cite{velickovic2017graph} (two commonly used general-purpose GNNs), as well as the current SOTA GNN model SimpleHGN~\cite{lv2021we}.
\end{itemize}

The configurations of baselines can be seen in Appendix~\ref{ss:conf}.
As a generic framework, \autoac{} can integrate different GNNs.
We select two representative GNN models from the two categories (i.e., MAGNN and SimpleHGN) from the perspective of performance and computational efficiency.
Then, we combine \autoac{} with the two models, denoted by MAGNN-\autoac{} and SimpleHGN-\autoac{} respectively.
%

Table~\ref{tbl:compare_gnn} shows the performance comparison between \autoac{} and existing heterogeneous GNNs on node classification. 
\autoac{} can improve the performance of MAGNN and SimpleHGN stably on all datasets. 
The performance gain obtained by \autoac{} over MAGNN is around 0.7\%-3\% and the error rate is reduced by 2.87\%-11.69\%.
Also, SimpleHGN-\autoac{} outperforms SimpleHGN by 1\%-3\% and reduces the error rate by 1.59\%-22.09\%.
By combining with the SOTA model SimpleHGN, SimpleHGN-\autoac{} can achieve the best performance in all models.

Moreover, Table~\ref{tbl:compare_gnn} shows that \autoac{} can bring significant performance improvement on the datasets where the classification target nodes have no raw attributes (e.g., DBLP).
Besides, for the datasets where the target nodes already have raw attributes (e.g., ACM and IMDB), completing other non-target nodes using \autoac{} can still promote the classification accuracy of target nodes.
Especially, for the IMDB dataset, since there are too many non-target nodes with missing attributes (i.e., 77\% of all nodes), the performance improvement with \autoac{} is more significant.

Note that the performance of MAGNN without attribute completion is not as good as other models, such as GTN and GAT.
However, MAGNN-\autoac{} performs better than GTN on DBLP and ACM, and outperforms GAT on DBLP and IMDB, which indicates that effective attribute completion for heterogeneous graphs can compensate for the performance gap introduced by the GNN model.
\textcolor{ black}{By unifying attribute completion and representation learning in an unsupervised heterogeneous network, the recently proposed HGCA can also achieve competitive performance on DBLP and ACM.}
Such experimental results further verify the necessity of \autoac{}.

\subsubsection{Performance comparison with the existing attribute completion method HGNN-AC}
As the current SOTA attribute completion method, HGNN-AC~\cite{jin2021heterogeneous} uses the attention mechanism to aggregate the attributes of the direct neighbors for the nodes with missing attributes.
The attention information is calculated by the pre-learning of topological embedding.
To be fair, both \autoac{} and HGNN-AC are evaluated under the unified HGB benchmark. 
And, we also combine HGNN-AC with MAGNN and SimpleHGN, denoted by MAGNN-HGNNAC and SimpleHGN-HGNNAC respectively.
%
%


Table~\ref{tbl:compare_hgnnac} shows that \autoac{} outperforms HGNN-AC on all datasets.
Specifically, MAGNN-\autoac{} achieves 1\%-4\% performance improvement over MAGNN-HGNNAC.
For the SimpleHGN model, SimpleHGN-\autoac{} outperforms SimpleHGN-HGNNAC by 0.4\%-2\%.
Moreover, the performance improvement of HGNN-AC for attribute completion is not stable.
As shown in Table~\ref{tbl:compare_hgnnac}, after attribute completion with HGNN-AC, MAGNN-HGNNAC is instead inferior to MAGNN on the three datasets, while MAGNN-\autoac{} can achieve significant performance improvement with attribute completion. 
Similarly, there is a degradation in performance on the DBLP dataset compared to SimpleHGN.
\subsubsection{Performance comparison on link prediction}
To verify the effectiveness of \autoac{} on different downstream tasks, we further conduct link prediction in Table~\ref{link prediction}. 
\autoac{} can greatly improve the performance of heterogeneous GNNs, especially on IMDB.
With \autoac{}, MRR and ROC-AUC of SimpleHGN are increased by 9.7\% and 28\%, respectively.

In summary, \autoac{} achieves better performance and more stable performance improvement, indicating the effectiveness of searching for the most suitable attribute completion operations for no-attribute nodes from a diverse search space.

\subsection{Efficiency Study}
Besides the effectiveness, we also evaluate the efficiency of \autoac{} in the terms of runtime overhead.
\textcolor{ black}{Table \ref{tbl:compare_gnn} and \ref{link prediction} show the runtime of \autoac{} and other handcrafted HGNNs on node classification and link prediction tasks.
Although the attribute completion and GNN training are jointly optimized in \autoac{}, the computational efficiency of \autoac{} is still competitive compared to other baselines. 
}

Also, we compare \autoac{} with the existing attribute completion method HGNN-AC.
Table~\ref{tbl:time} shows the efficiency comparison between \autoac{} and HGNN-AC.
\autoac{} contains the search and retraining stages, and HGNN-AC contains the pre-learning and training stages.
We can see that \autoac{} is much more efficient than HGNN-AC.
The end-to-end runtime overhead of \autoac{} can be reduced by 15$\times$ to 465$\times$.
The main reason why HGNN-AC is inefficient is that the pre-leaning stage that learns a topological embedding for each node
is very time-consuming.
Especially for the DBLP dataset with a large number of nodes, the pre-learning overhead is up to 9 GPU hours.
In contrast, there is no additional pre-leaning stage in \autoac{}. 
Moreover, by introducing the discrete constraints and auxiliary unsupervised clustering task, the search efficiency can be improved significantly.

In summary, 
\autoac{}
can not only achieve better performance but also demonstrate higher computational efficiency.

\subsection{Ablation Study}

\begin{table*}
  \caption{Completion operation ablation study on SimpleHGN. Bold indicates the global best. Underline indicates the best among all single attribute completion operations.}
  \centering
  \scalebox{0.85}{
  \label{Operator Ablation experiments on SimpleHGN}
  
  \begin{tabular}{l|cc|cc|cc}
    \toprule
    Dataset   & \multicolumn{2}{c}{DBLP}  & \multicolumn{2}{c}{ACM}  & \multicolumn{2}{c}{IMDB} \\
    \midrule
    Model $\backslash$ Metrics & Macro-F1 & Micro-F1 & Macro-F1 & Micro-F1 & Macro-F1 & Micro-F1 \\
    \midrule
    Baseline (SimpleHGN) & 93.83$\pm$0.18	&94.25$\pm$0.19&	92.92$\pm$0.67&	92.85$\pm$0.68&	62.98$\pm$1.66&	67.42$\pm$0.42 \\
    \midrule
    GCN\_AC & \underline{94.23$\pm$0.21} &	\underline{94.88$\pm$0.23}&	93.25$\pm$0.45&	93.18$\pm$0.47&	\underline{64.67$\pm$0.94}&	\underline{67.96$\pm$0.53} \\
    PPNP\_AC & 85.76$\pm$2.24&	86.58$\pm$2.23&	\underline{93.42$\pm$0.46}&	\underline{93.34$\pm$0.48}&	53.36$\pm$19.31	&61.68$\pm$11.76 \\
   MEAN\_AC & 90.91$\pm$0.72	&91.53$\pm$0.67&	92.99$\pm$0.60&	92.90$\pm$0.62&	63.73$\pm$0.94&	67.61$\pm$0.30\\
   One-hot\_AC &93.80$\pm$0.13&	94.30$\pm$0.14&	93.38$\pm$0.16&	93.31$\pm$0.15&	64.17$\pm$0.83&	67.89$\pm$0.24 \\
   \midrule
   Random\_AC &91.28$\pm$1.63&	91.77$\pm$1.55&	93.02$\pm$0.29&	92.95$\pm$0.31&	64.03$\pm$0.68&	67.43$\pm$0.33\\
   \midrule
   \textbf{\autoac{}} & \textbf{95.15$\pm$0.29}&	\textbf{95.52$\pm$0.26}&	\textbf{93.86$\pm$0.18}&	\textbf{93.80$\pm$0.18}&	\textbf{64.92$\pm$0.58}&	\textbf{67.94$\pm$0.41}\\
    \bottomrule
  \end{tabular}}
\end{table*}

\begin{table*}
  \caption{Completion operation ablation study on MAGNN. Bold indicates the global best. Underline indicates the best among all single attribute completion operations.}
   \centering
   \scalebox{0.85}{
  \label{Operator Ablation experiments on MAGNN}
  \begin{tabular}{l|cc|cc|cc}
    \toprule
    Dataset   & \multicolumn{2}{c}{DBLP}  & \multicolumn{2}{c}{ACM}  & \multicolumn{2}{c}{IMDB} \\
    \midrule
    Model $\backslash$ Metrics & Macro-F1 & Micro-F1 & Macro-F1 & Micro-F1 & Macro-F1 & Micro-F1 \\
    \midrule
    Baseline (MAGNN) & 93.16$\pm$0.38&	93.65$\pm$0.34&	91.06$\pm$1.44&	90.95$\pm$1.43&	56.92$\pm$1.76&	65.11±0.59 \\
    \midrule
    GCN\_AC & 93.74$\pm$0.34&	94.16$\pm$0.34&	90.96$\pm$0.77&	90.87$\pm$0.76&	57.96$\pm$1.11&	65.71$\pm$0.50 \\
    PPNP\_AC & 93.46$\pm$0.32&	93.94$\pm$0.29&	90.38$\pm$0.67&	90.28$\pm$0.67&	\underline{58.46$\pm$1.17}&	\underline{65.97$\pm$0.56} \\
   MEAN\_AC & \underline{93.89$\pm$0.12}&\underline{94.33$\pm$0.13}&90.97$\pm$0.48&	90.86$\pm$0.49&	57.60$\pm$0.71&	65.42$\pm$0.38\\
   One-hot\_AC &93.73$\pm$0.32&	94.15$\pm$0.28&	\underline{91.04$\pm$0.69}&	\underline{90.92$\pm$0.70}&	58.12$\pm$1.71&	65.43±0.68 \\
   \midrule
   Random\_AC &93.38$\pm$0.25&	93.87$\pm$0.19&	91.09$\pm$0.61&	90.98$\pm$0.63&	57.97$\pm$1.15&	65.57$\pm$0.77\\
   \midrule
   \textbf{\autoac{}} & \textbf{93.95$\pm$0.30}&	\textbf{94.39$\pm$0.25}&	\textbf{91.84$\pm$0.45}&	\textbf{91.77$\pm$0.45}&	\textbf{58.96$\pm$1.31}&	\textbf{66.11$\pm$0.53}\\
    \bottomrule
  \end{tabular}}
  \vspace{-2ex}
\end{table*}

\subsubsection{Study on the necessity of searching attribute completion operations from a diverse search space}
%
We compare \autoac{} with the following two methods: 
\begin{itemize}
\item \textbf{Single-operation attribute completion:} We complete all no-attribute nodes with the same single completion operation (i.e., GCN$\_$AC, PPNP$\_$AC, MEAN$\_$AC, and One-hot$\_$AC). 
\item \textbf{Random attribute completion:} For each no-attribute node, we randomly select an attribute completion operation from the search space.
\end{itemize}

%
Table~\ref{Operator Ablation experiments on SimpleHGN} and Table~\ref{Operator Ablation experiments on MAGNN} show the completion operation ablation study on SimpleHGN and MAGNN.
Due to the differences in the data characteristics, there is no single completion operation that can perform well on all datasets. 
By searching the optimal attribute completion operations
\autoac{} can achieve the best performance on all datasets.

Take SimpleHGN shown in Table~\ref{Operator Ablation experiments on SimpleHGN} for example. 
GCN\_AC is more effective on DBLP and IMDB, while PPNP\_AC performs better on ACM. 
Moreover, for a specific attribute completion operation, the performance is related to the dataset and the chosen GNN model.
We take DBLP as an example.
GCN\_AC performs better on SimpleHGN.
However, when the GNN model becomes MAGNN, GCN\_AC is not as good as MEAN\_AC.
Additionally, the performance of the random attribute completion is not stable and can be even worse than the baseline model.
Choosing an inappropriate completion operation can have a negative effect on the final performance.


\begin{table*}
  \caption{Ablation study on discrete constraints. $/$ indicates memory overflow.} 
\centering
  \scalebox{0.85}{
  \label{Proximal Iterative Ablation in MAGNN and SimpleHGN}
  \begin{tabular}{l|ccc|ccc|ccc}
    \toprule
    Dataset   & \multicolumn{3}{c}{DBLP}  & \multicolumn{3}{c}{ACM}  & \multicolumn{3}{c}{IMDB} \\
    \midrule
    Model $\backslash$ Metrics & Macro-F1 & Micro-F1 & \makecell{Search Time \\ (Seconds)} & Macro-F1 & Micro-F1&  \makecell{Search Time \\ (Seconds)} & Macro-F1 & Micro-F1 &  \makecell{Search Time \\ (Seconds)} \\
    \midrule
    \textbf{SimpleHGN-\autoac{}} & \textbf{95.15$\pm$0.29}&	\textbf{95.52$\pm$0.26}&\textbf{32}&\textbf{93.86$\pm$0.18}	&\textbf{93.80$\pm$0.18}&\textbf{72}&\textbf{64.92$\pm$0.58}&	\textbf{67.94$\pm$0.41} &\textbf{36}\\
   w/o Discrete constraints &95.12$\pm$0.27&	95.49$\pm$0.25& 216&	93.43$\pm$0.74&	93.34$\pm$0.76& 360&	64.74$\pm$0.68&	67.85$\pm$0.52& 180\\
   \midrule
   \textbf{MAGNN-\autoac{}} & \textbf{93.95$\pm$0.30}&	\textbf{94.39$\pm$0.25}&\textbf{72}&	\textbf{91.84$\pm$0.45}	&\textbf{91.77$\pm$0.45}&\textbf{432}&\textbf{58.96$\pm$1.31}&	\textbf{66.11$\pm$0.53}&\textbf{504} \\
   w/o Discrete constraints &	/ &	/&/&	91.24$\pm$0.67&	91.45$\pm$0.68& 1800 &	58.44$\pm$1.12&	65.65$\pm$0.34& 1908 \\
    \bottomrule
  \end{tabular}}
  \vspace{-3ex}
\end{table*}

\subsubsection{Study on the search algorithm with discrete constraints}
When optimizing the attribute completion parameters $\alpha$, we enforce discrete constraints on $\alpha$ and solve the bi-level optimization problem with proximal iteration. 
To verify the effectiveness of discrete constraints, we further run \autoac{} with and without discrete constraints in Table~\ref{Proximal Iterative Ablation in MAGNN and SimpleHGN}.

%
The search algorithm with discrete constraints can achieve better performance with less search time overhead on all datasets. 
%
%
%
%
Additionally, proximal iteration allows removing the need for second-order derivative in solving the bi-level optimization problem. 
Thus, the memory overhead can also be reduced significantly.
As shown in Table~\ref{Proximal Iterative Ablation in MAGNN and SimpleHGN}, the memory overhead of MAGNN-\autoac{} without discrete constraints is huge and the out-of-memory error occurs on DBLP.

\begin{figure}
  \centering
      \subfigure[SimpleHGN]{
    \includegraphics[width=0.9\columnwidth]{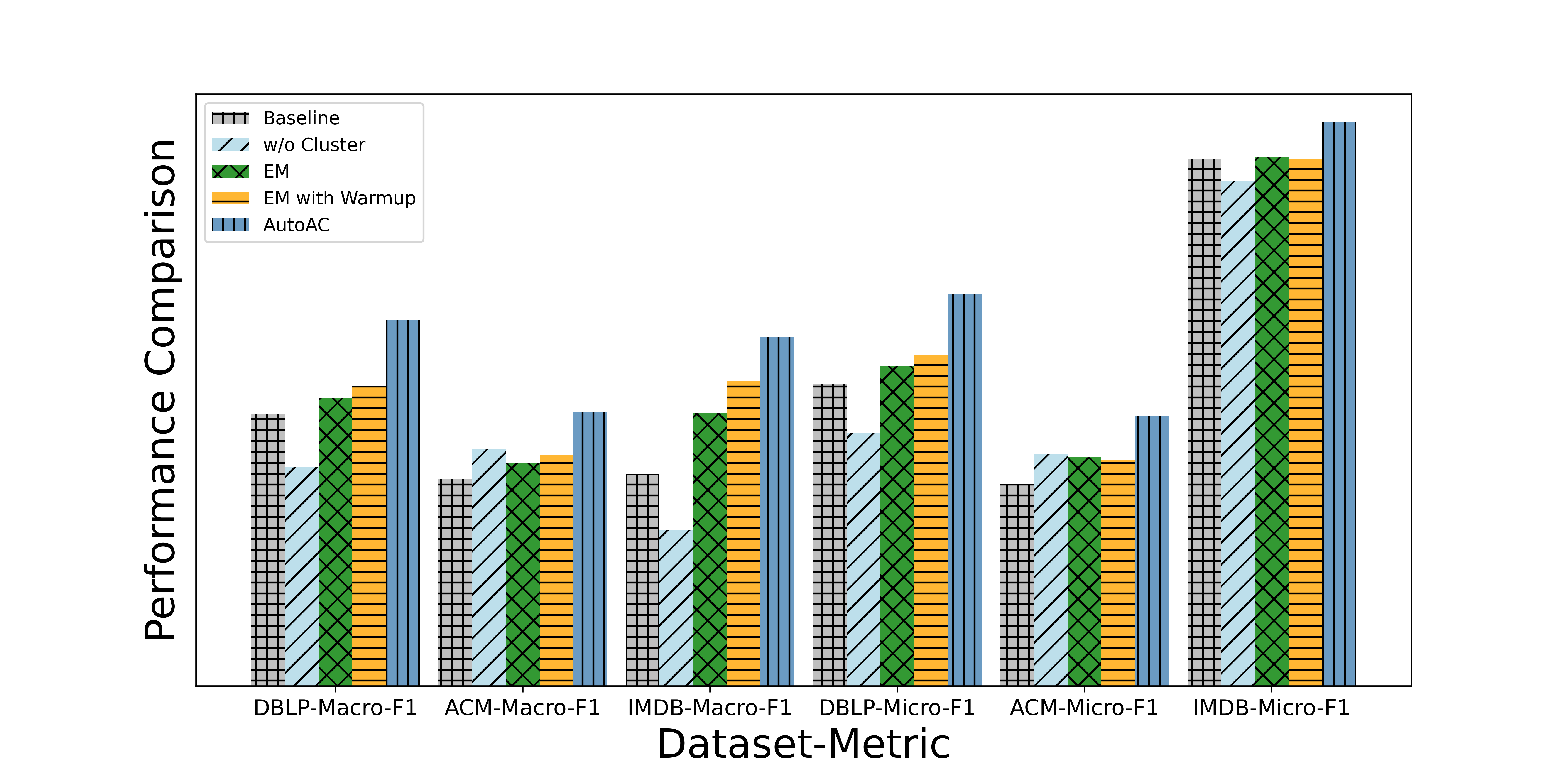}
    }
    \subfigure[MAGNN]{
\includegraphics[width=0.9\columnwidth]{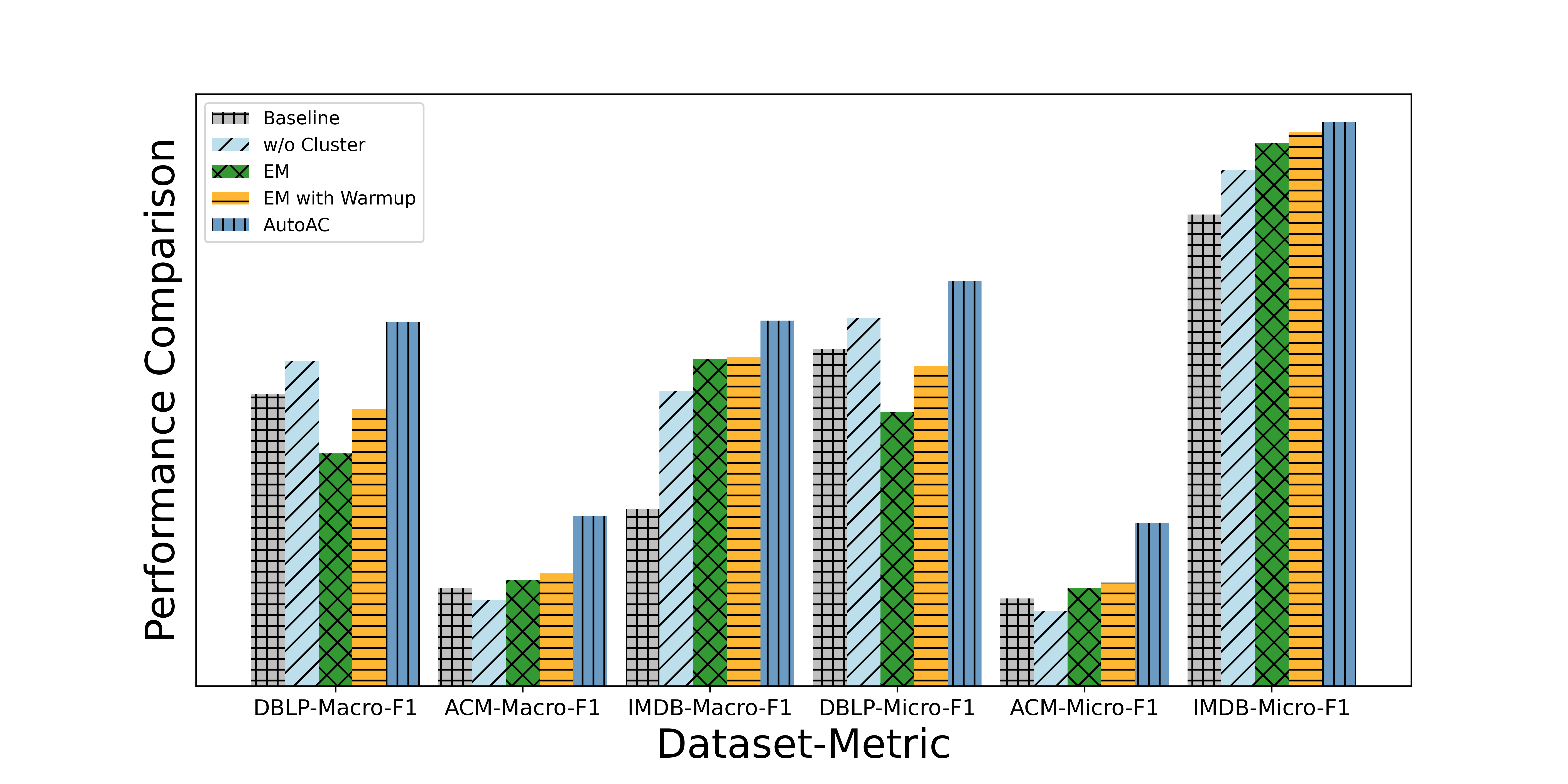}
    }
    \vspace{-1ex}
  \caption{Performance comparison between different clustering methods.}
  \label{Performance comparison of using different clusters optimization algorithms}
  \vspace{-2ex}
\end{figure}

\begin{figure}[t]
  \centering
  \subfigure[DBLP]{
    \includegraphics[width=0.29\columnwidth]{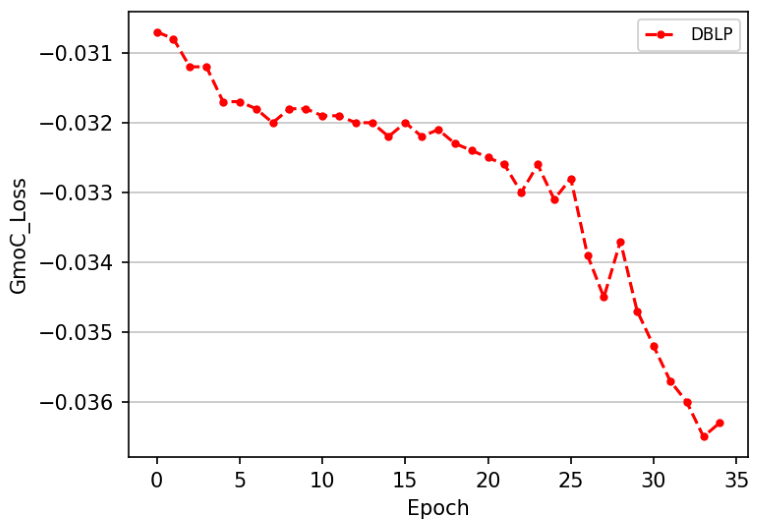}
    }
      \subfigure[ACM]{
    \includegraphics[width=0.29\columnwidth]{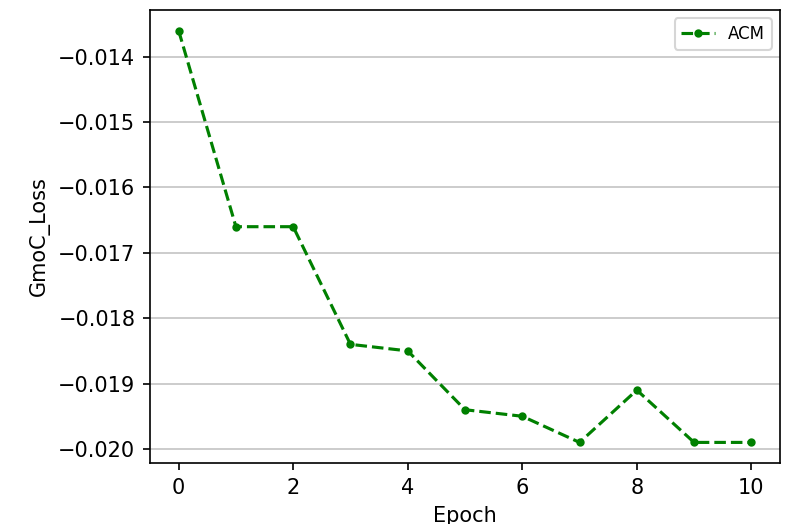}
    }
    \subfigure[IMDB]{
    \includegraphics[width=0.29\columnwidth]{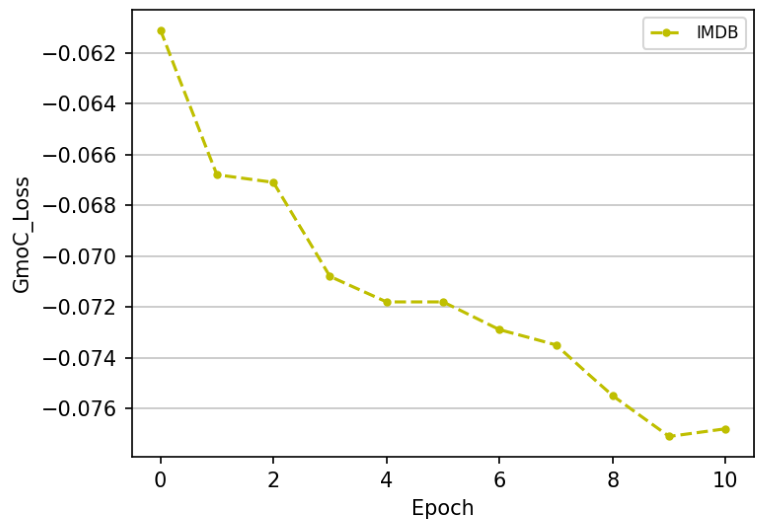}
    }
  \caption{Convergence of $\mathcal{L}_{G m o C}$ on three datasets.}
  \vspace{-3ex}
  \label{Convergence on three datasets}
\end{figure}

\subsubsection{Study on the auxiliary unsupervised clustering}

To reduce the dimension of the completion parameters $\alpha$, we leverage 
an auxiliary unsupervised clustering task.
%
Figure~\ref{Performance comparison of using different clusters optimization algorithms} shows the performances of different clustering methods.
\begin{itemize}
\item \textbf{w/o cluster:} We directly search the attribute completion operations for each no-attribute node without clustering. 
\item \textbf{EM:} After each iteration of the optimization process, we adopt the EM algorithm for clustering according to node representation learned by the GNN model.
\item \textbf{EM with warmup:} a variant of the EM algorithm, which adds a warm-up process at the beginning of the clustering.
\end{itemize}

\begin{figure}[t]
  \centering
    \includegraphics[width=1.0\columnwidth]{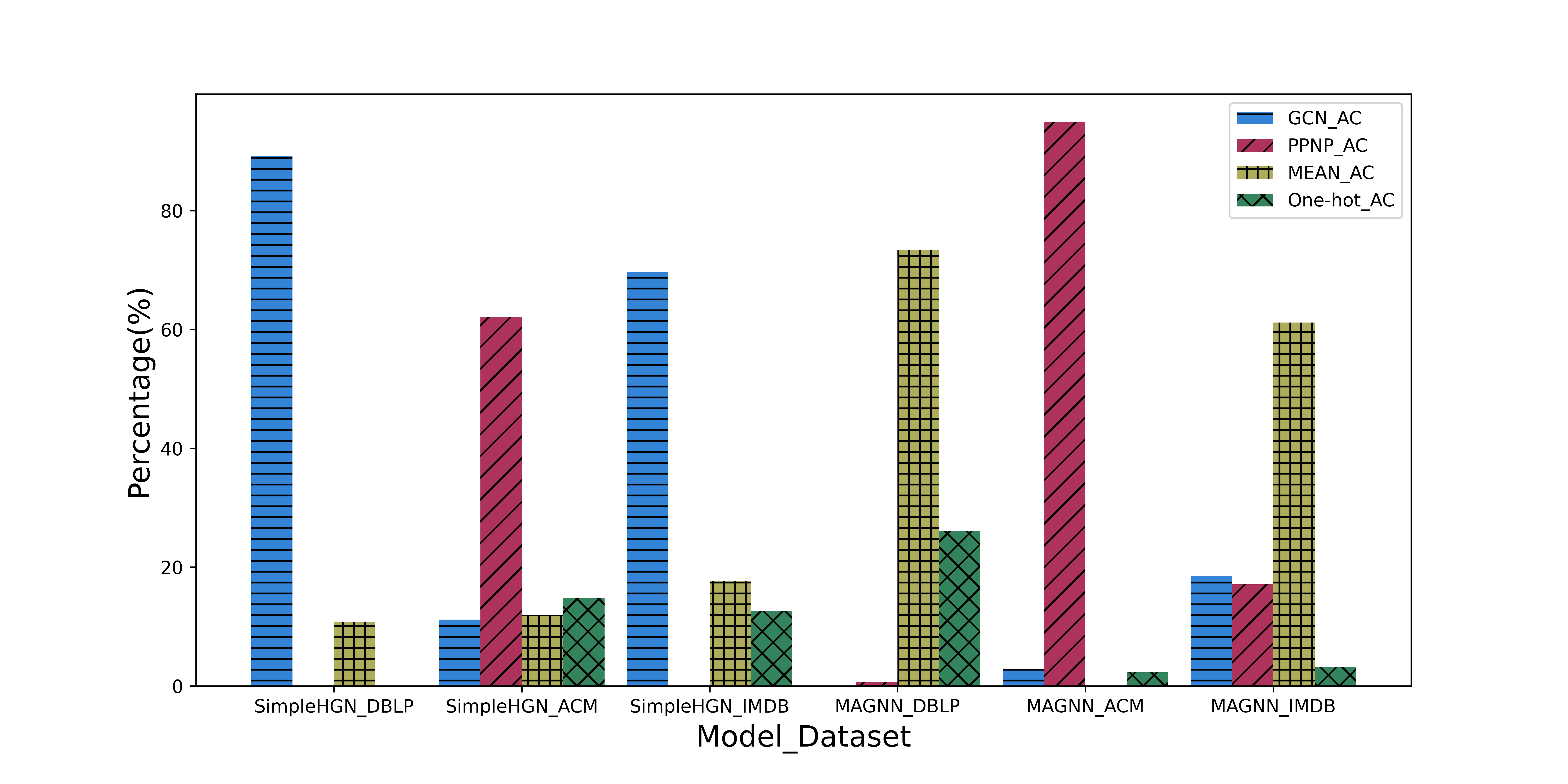}
    \vspace{-6ex}
  \caption{Distribution of searched attribute completion operations.}
  \label{Attribute completions selected on different models}
  \vspace{-1ex}
\end{figure}

\begin{figure}
  \centering
      \subfigure[Author]{
    \includegraphics[width=0.25\columnwidth]{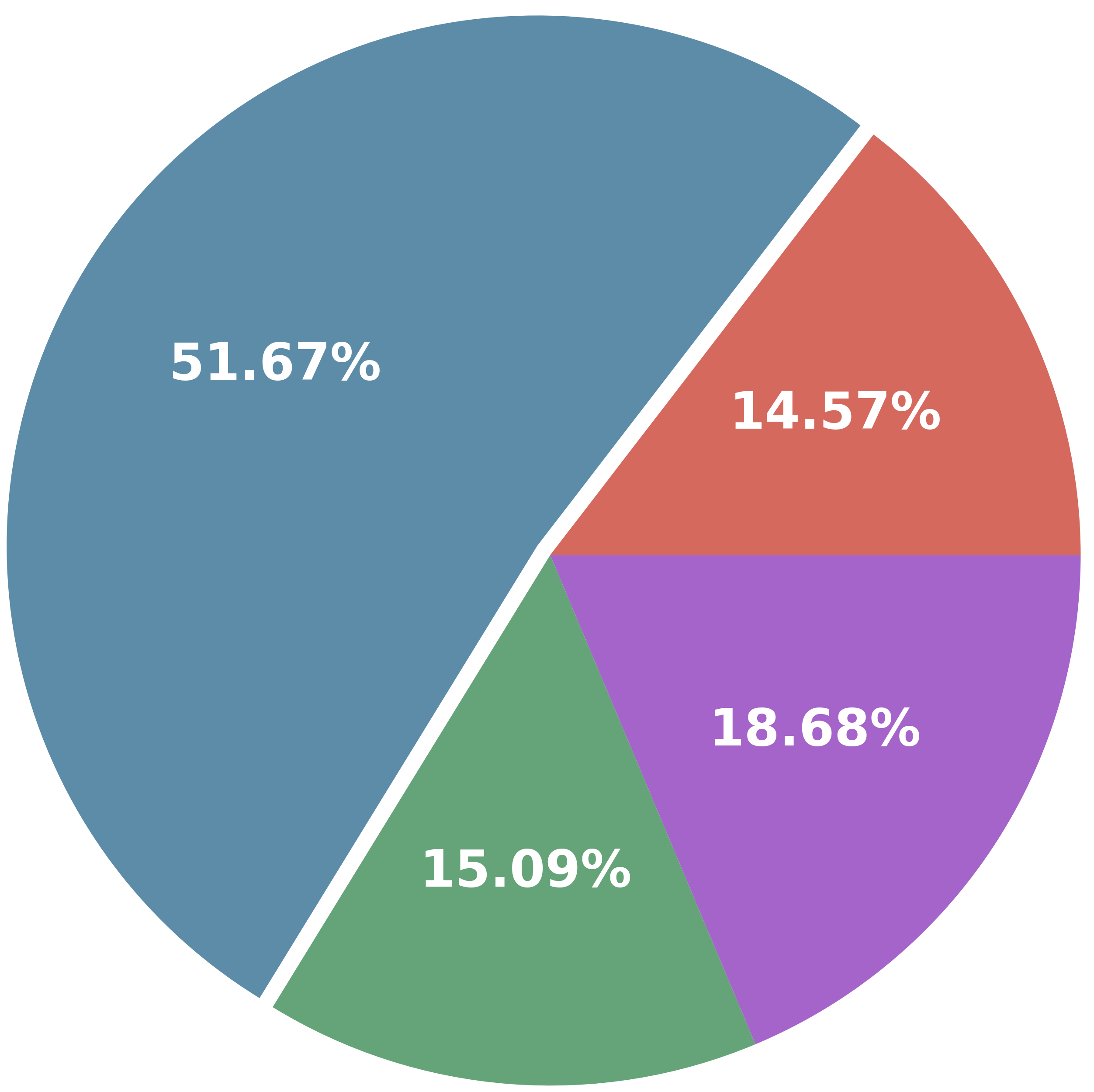}
    }
    \hspace{0.0005\textwidth}
    \subfigure[Subject]{
    \includegraphics[width=0.25\columnwidth]{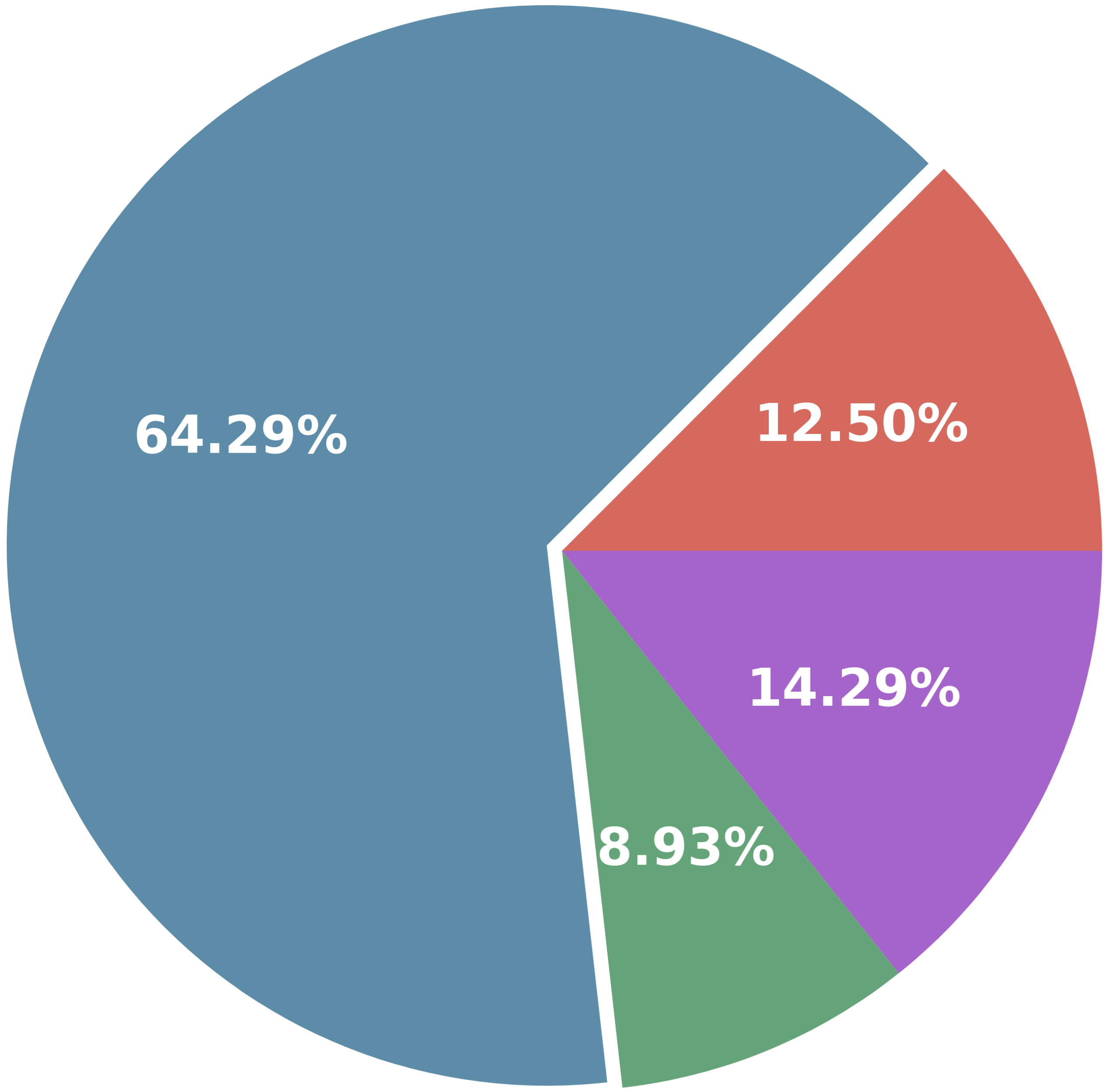}
    }
    \hspace{0.0005\textwidth}
    \subfigure[Term]{
    \includegraphics[width=0.3\columnwidth]{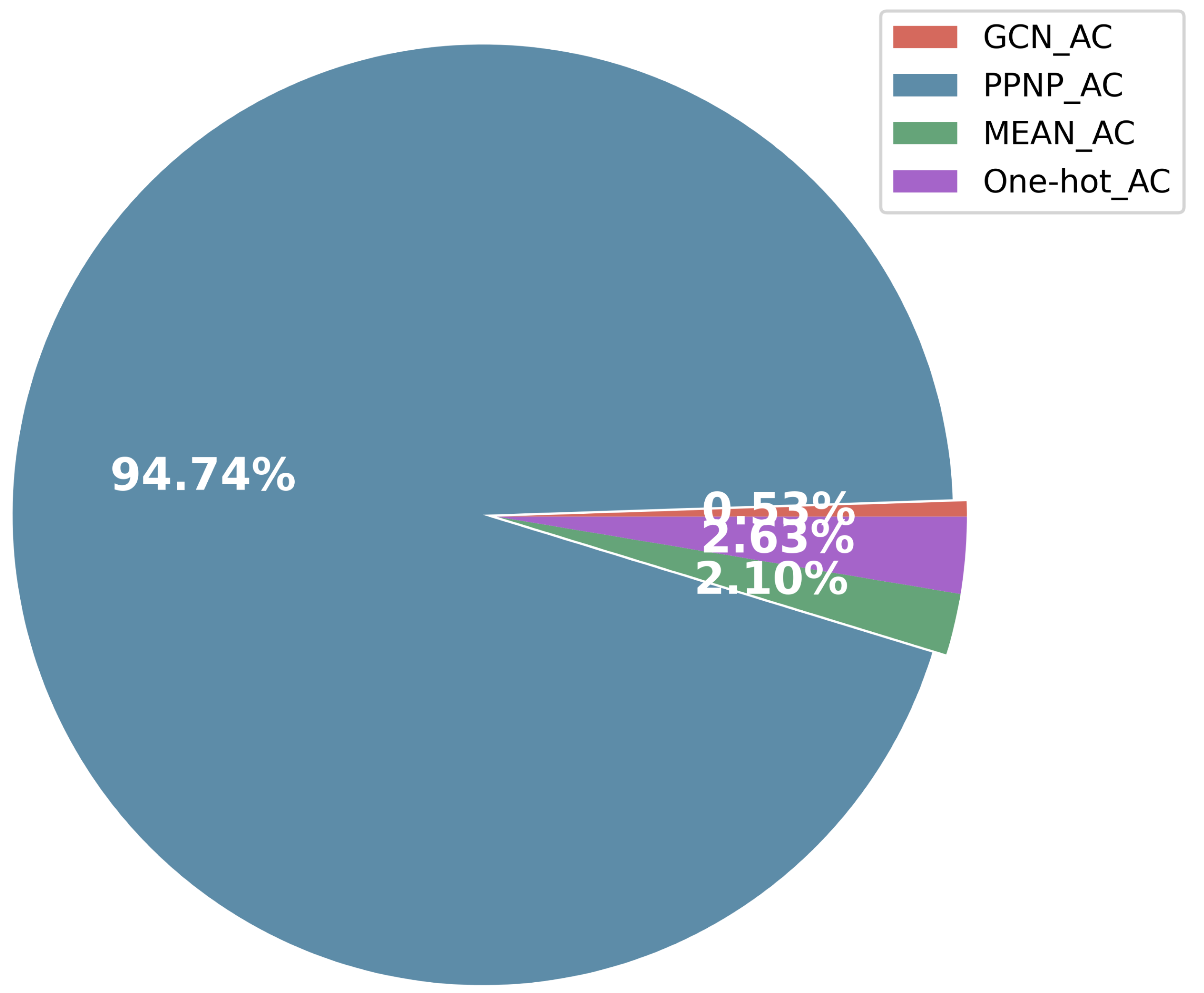}
    }
    \vspace{-1ex}
      \caption{Detailed distribution of searched completion operations for each no-attribute node type on the ACM dataset using SimpleHGN-\autoac{}.}
  \label{ACM_example}
  \vspace{-3ex}
\end{figure}

\begin{figure}[th]
  \centering
      \subfigure[Actor]{
    \includegraphics[width=0.25\columnwidth]{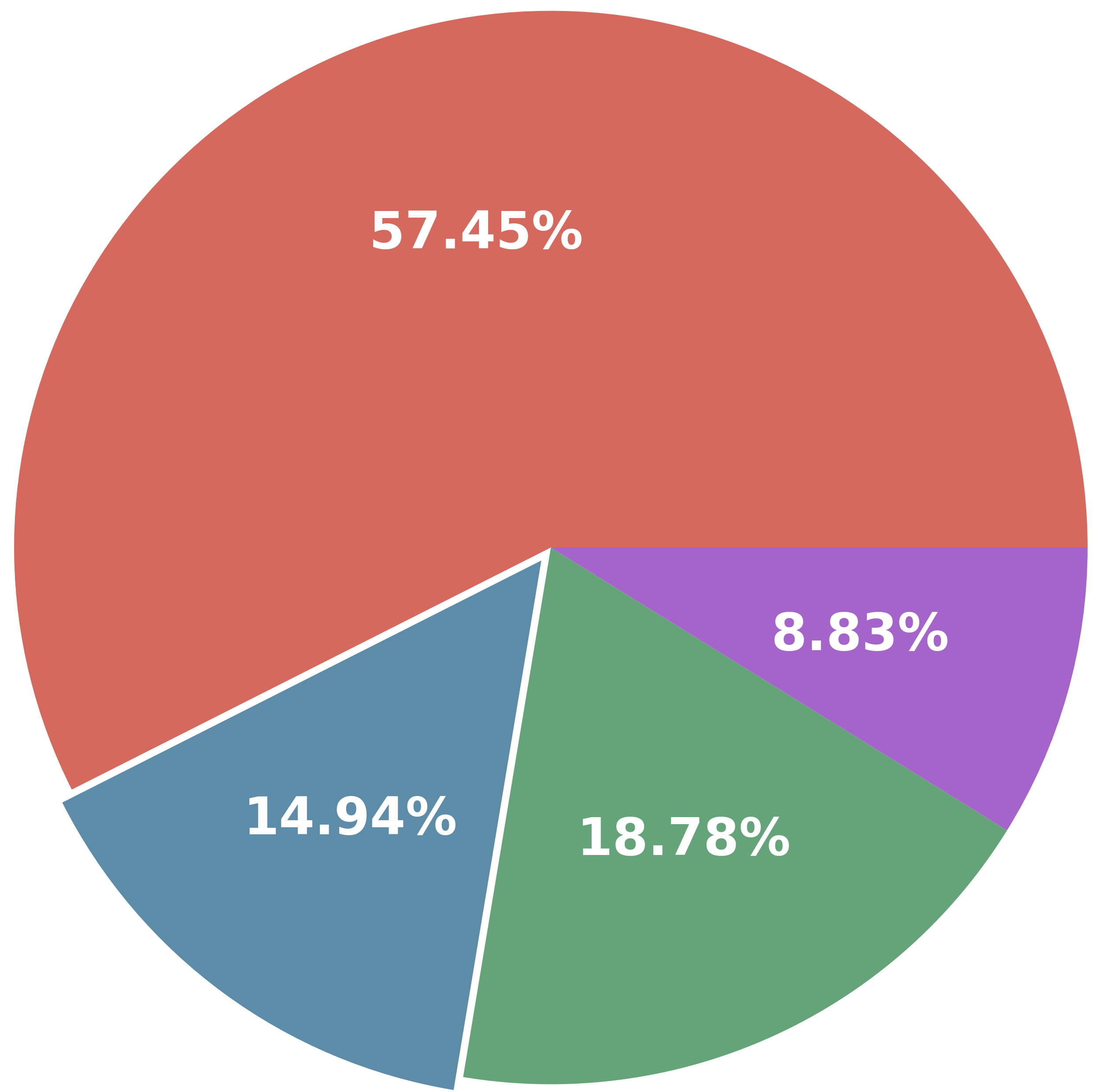}
    }
    \hspace{0.0005\textwidth}
    \subfigure[Director]{
    \includegraphics[width=0.25\columnwidth]{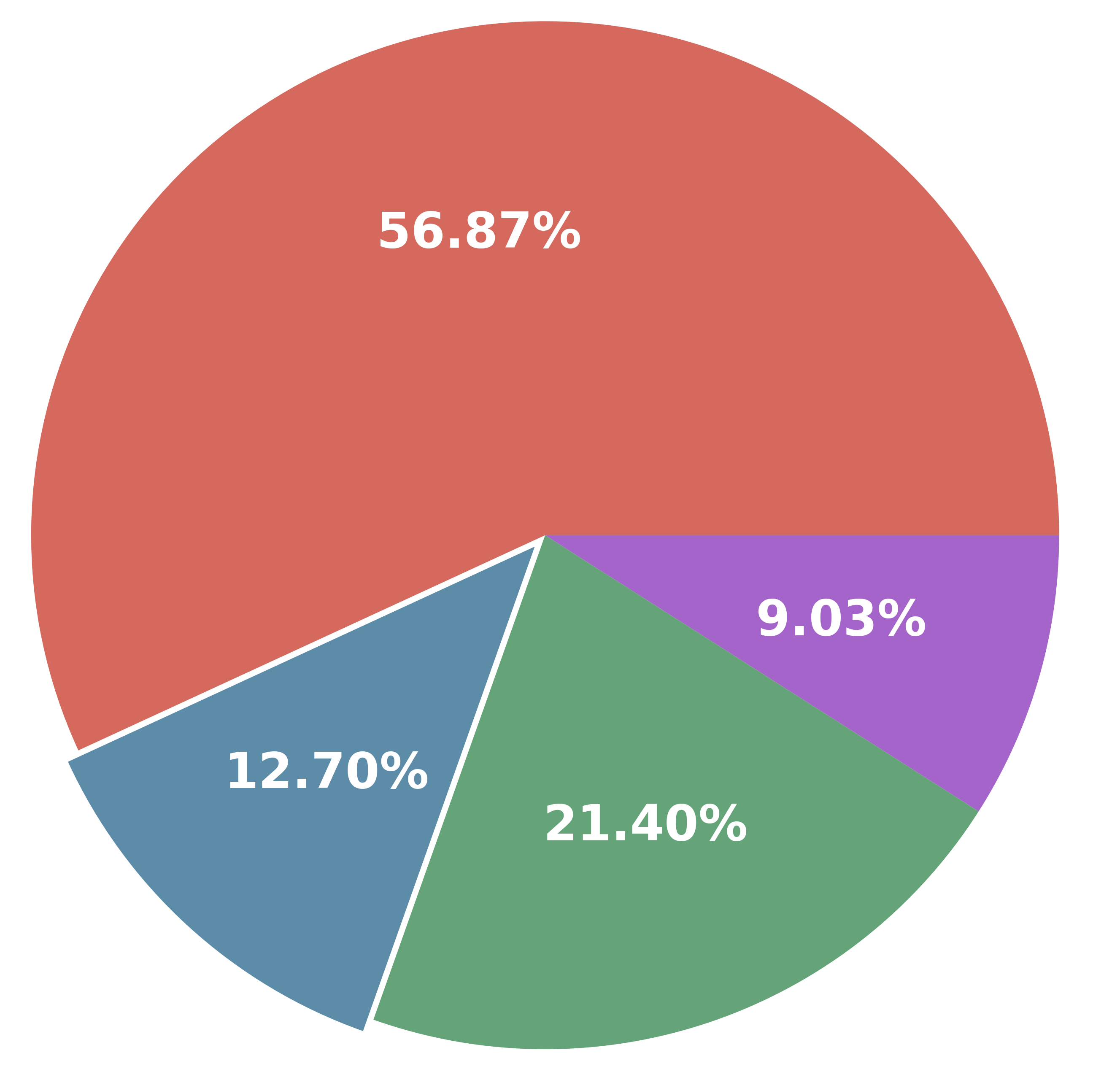}
    }
    \hspace{0.0005\textwidth}
    \subfigure[Keyword]{
    \includegraphics[width=0.3\columnwidth]{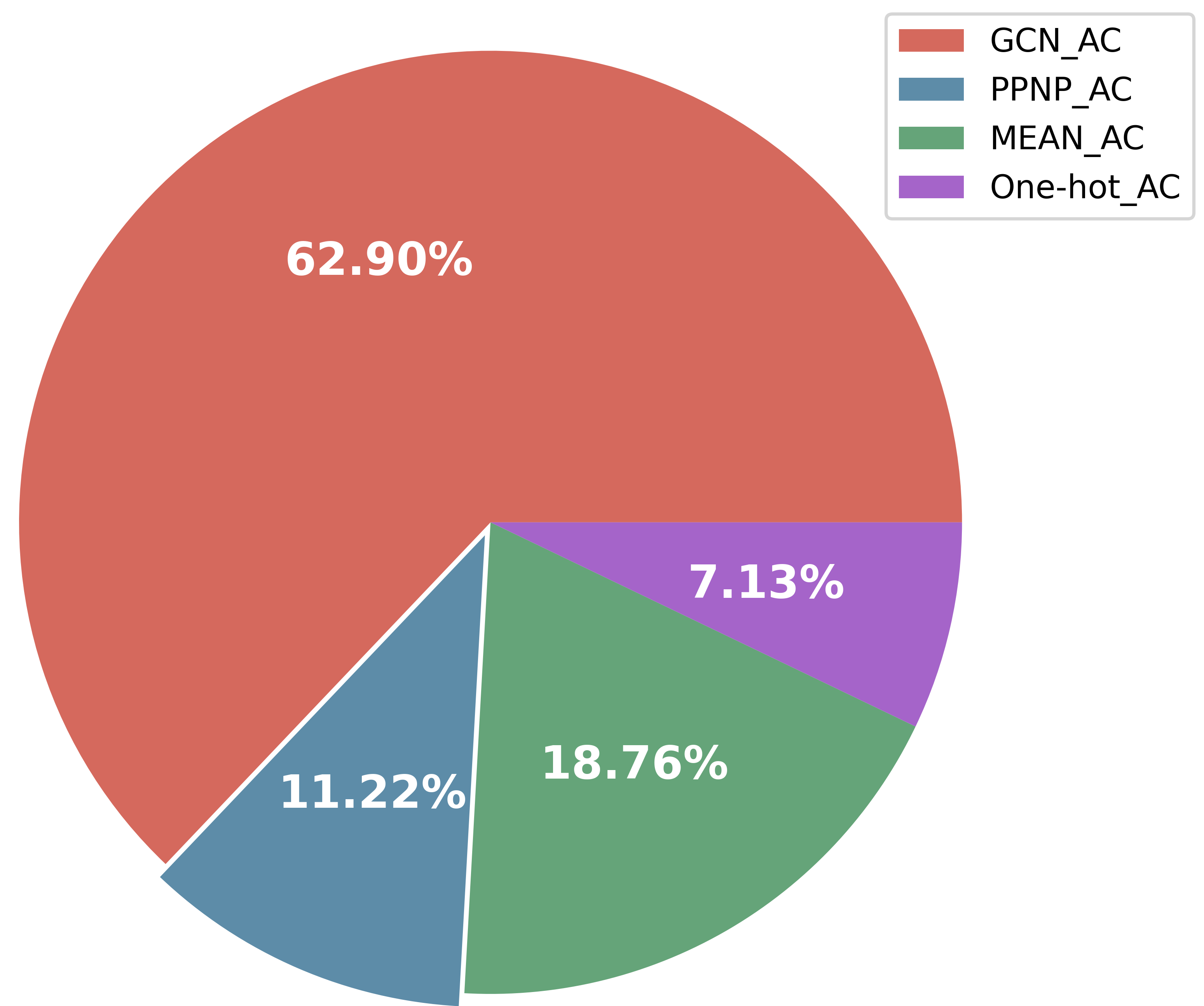}
    }
    \vspace{-1ex}
  \caption{\textcolor{ black}{Detailed distribution of searched completion operations for each no-attribute node type on the IMDB dataset using SimpleHGN-\autoac{}.}}
  \label{IMDB_example}
  \vspace{-2ex}
\end{figure}

In Figure~\ref{Performance comparison of using different clusters optimization algorithms}, \autoac{} can achieve the best performance on all datasets. 
Searching completion operations without clustering yields relatively poor performance.
%
%
%
Reducing the dimension of $\alpha$ with unsupervised clustering is very necessary. 
Moreover, the proposed unsupervised clustering method outperforms EM and its variant, indicating the effectiveness of the joint optimization of the unsupervised clustering loss and the classification loss.
Figure~\ref{Convergence on three datasets} also shows the convergence of the unsupervised clustering loss $\mathcal{L}_{G m o C}$, which exhibits a stable decreasing trend during the optimization process.

\subsection{Distribution of Searched Completion Operations}
Figure~\ref{Attribute completions selected on different models} shows the proportion of attribute completion operations searched by SimpleHGN-\autoac{} and MAGNN-\autoac{}.
For different models and datasets, the proportions of searched completion operations are quite different. 
In SimpleHGN-\autoac{}, DBLP tends to select GCN\_AC, while ACM prefers PPNP\_AC.
For the same dataset, different GNNs also result in different distributions. 
Take DBLP as an example. MAGNN-\autoac{} is more inclined to MEAN\_AC than GCN\_AC compared to SimpleHGN-\autoac{}. 
The results further indicate the necessity of searching for suitable attribute completion operations under different datasets and GNNs. 

\textcolor{ black}{
Figure~\ref{ACM_example} and Figure~\ref{IMDB_example} show the proportion of searched completion operations for each no-attribute node type on ACM and IMDB.
For ACM, multiple different completion operations are selected even for the same node type. 
Specifically, more than half of the author and subject nodes choose PPNP\_AC, while the proportions of other three operations are quite similar.
Most term nodes are assigned PPNP\_AC (i.e., 94.74\%), indicating that the term type is more likely to capture the global information.
%
%
%
The main reason is that the target node type (i.e., paper) with raw attributes in ACM contains only the paper title.
The high-order PPNP\_AC operations are preferred.
In contrast, GCN\_AC accounts for the majority of completion operations on IMDB. 
This is because the target node type (i.e., movie) has raw attributes and contains rich features, such as length, country, language, likes of movies, and ratings.
Thus, the local completion operation GCN\_AC is appropriate.}

\textcolor{ black}{
Next, we analyze the completion operations of concrete actor nodes. 
In IMDB, node No.10797 is the actor Leonardo DiCaprio, who has starred in 22 movies, and the neighborhood information is very rich. 
As a result, \autoac{} chooses GCN\_AC for him.
In contrast, node No.10799 is the actor Leonie Benesch, who has appeared in only one movie. 
Thus, one-hot\_AC is automatically selected by \autoac{}.
}
\subsection{Hyperparameter Sensitivity}
\label{ss:hyperparameters}
\subsubsection{
Effect of the number of clusters $M$}
\begin{figure}[t]
  \centering
    \subfigure[DBLP]{
    \includegraphics[width=0.29\columnwidth]{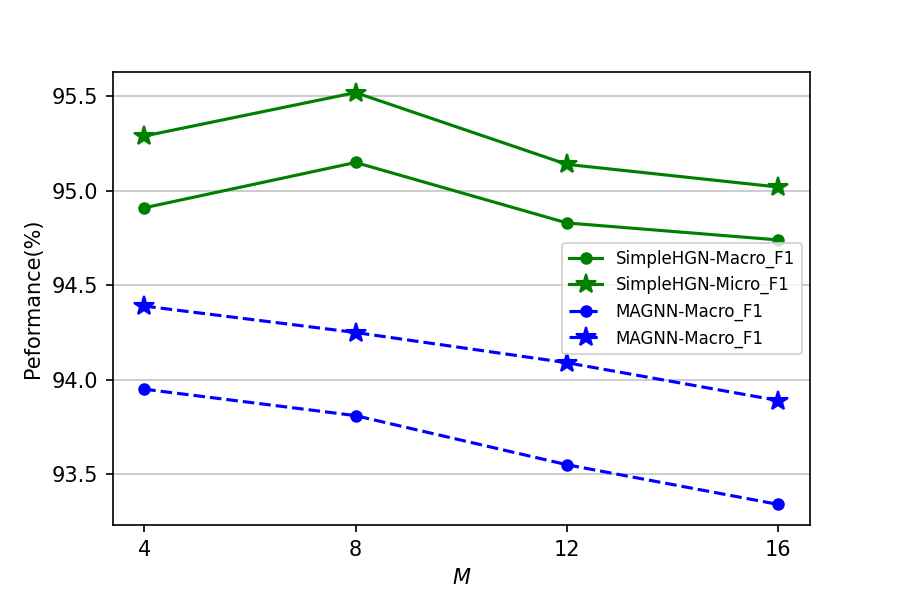}
    }
    \subfigure[ACM]{
    \includegraphics[width=0.29\columnwidth]{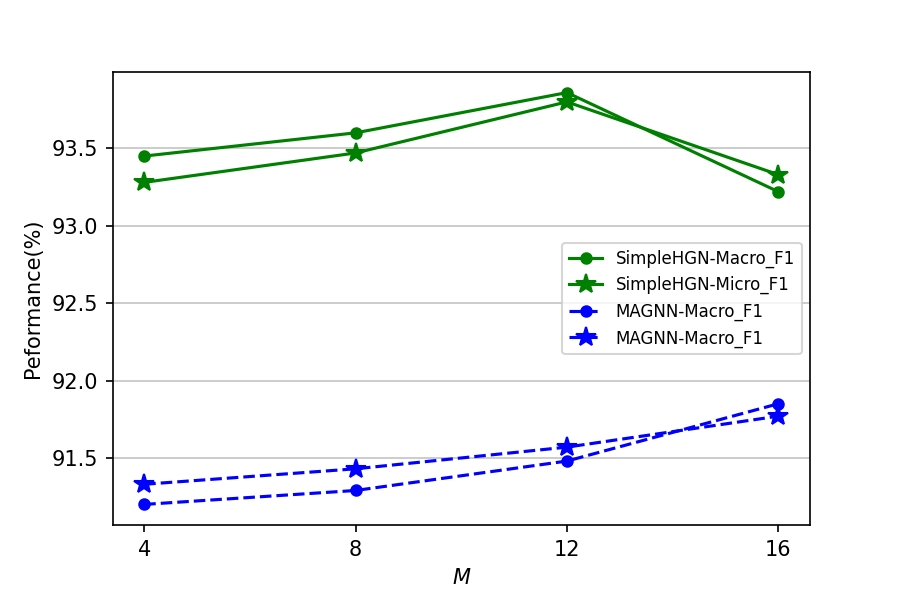}
    }
    \subfigure[IMDB]{
    \includegraphics[width=0.29\columnwidth]{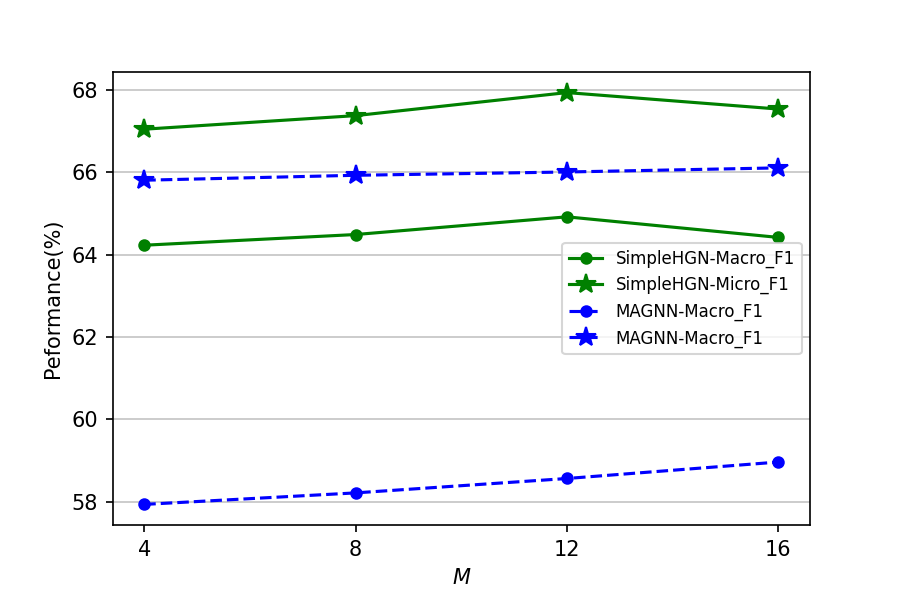}
    }
\vspace{-1ex}
  \caption{Performance comparison under different $M$.}
  \label{Performance comparison of AutoHC with different m}
\vspace{-3ex}
\end{figure}

Figure~\ref{Performance comparison of AutoHC with different m} shows the performance of \autoac{} under different $M$.
Both SimpleHGN-\autoac{} and MAGNN-\autoac{} can achieve stable performance, showing that \autoac{} has sufficient robustness to $M$.

\subsubsection{Effect of the loss weighted coefficient $\lambda$}
\begin{figure}[t]
  \centering
    \subfigure[DBLP]{
    \includegraphics[width=0.29\columnwidth]{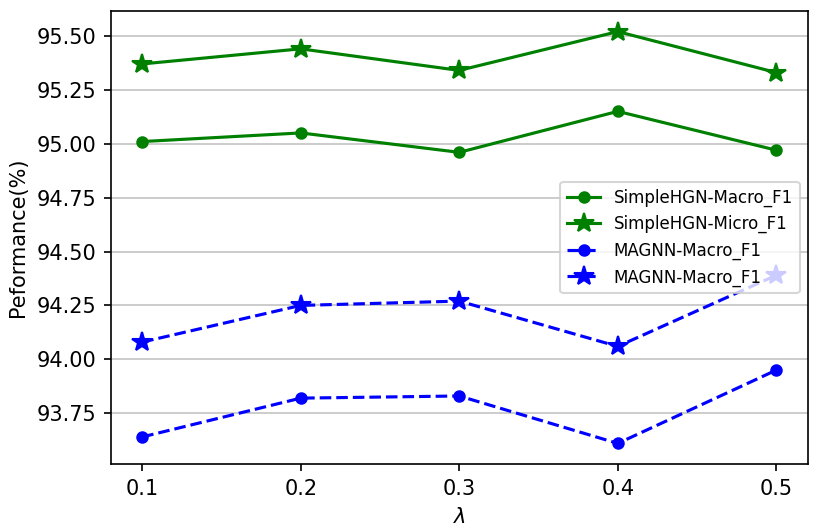}
    }
    \subfigure[ACM]{
    \includegraphics[width=0.29\columnwidth]{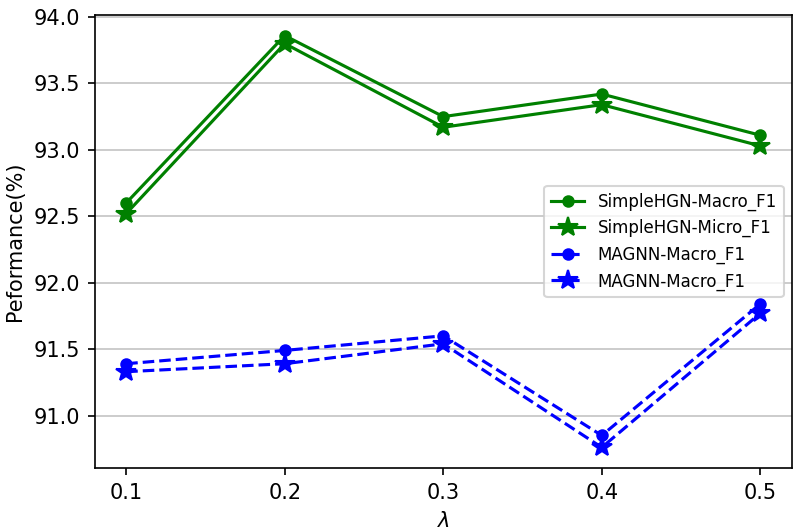}
    }
    \subfigure[IMDB]{
    \includegraphics[width=0.29\columnwidth]{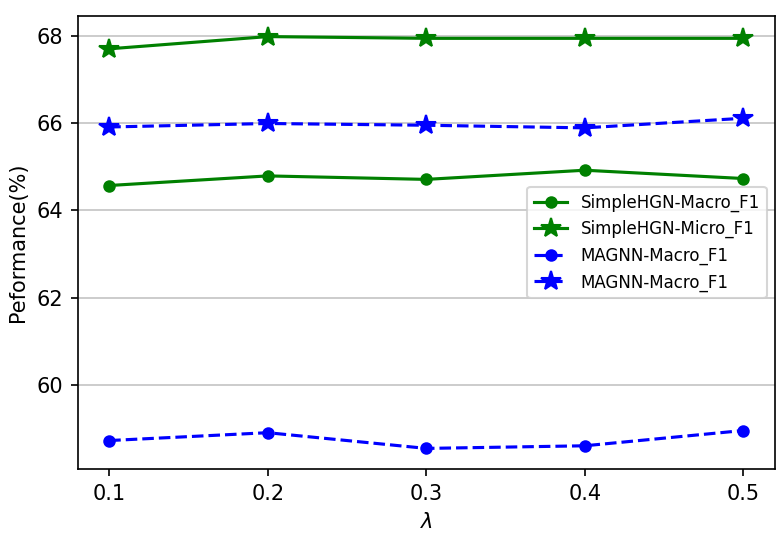}
    }
\vspace{-1ex}
  \caption{Performance comparison under different $\lambda$}
  \label{Performance comparison of AutoHC under different lamdba}
  \vspace{-2ex}
\end{figure}

We further evaluate the weighted coefficient $\lambda$ of the auxiliary unsupervised clustering loss.
The available values of $\lambda$ are set to [0.1, 0.2, 0.3, 0.4, 0.5].
Figure~\ref{Performance comparison of AutoHC under different lamdba} shows the performances of \autoac{} under different $\lambda$.
%
IMDB is very robust to $\lambda$, and the performance change is very insignificant.
For DBLP, $\lambda=0.4$ and $\lambda=0.5$ are suitable for SimpleHGN and MAGNN, respectively. 
For ACM, the choice of $\lambda$ is slightly sensitive.
%

The effects of the learning rate and the weight decay can be seen in Appendix~\ref{ss:lr}.

\begin{table}[t]
  \caption{\textcolor{ black}{Performance of SimpleHGN-\autoac{} with varying attribute missing rates in the node classification task.}} 
  \centering
  \scalebox{0.85}{
  \label{tbl:attrmiss}
  \begin{tabular}{ccccc}
    \toprule
    Datasets   & \makecell{Attribute \\ Missing Rates}  
   & \makecell{Node Types with \\ Missing attributes} &Macro-F1 &Micro-F1 \\
    \midrule
    \multirow{3}{*}{DBLP} & 0\% & / & 93.83$\pm$0.18& 94.25$\pm$0.19\\
    \multirow{3}{*}{} & 15\% &  author & 94.35$\pm$0.17& 94.72$\pm$0.16\\
    \multirow{3}{*}{} & 30\% &term, venue& 95.09$\pm$0.13& 95.47$\pm$0.12\\
    \multirow{3}{*}{} & 45\% &author, term, venue & \textbf{95.15$\pm$0.29} & \textbf{95.52$\pm$0.26} \\
    \midrule
    \multirow{3}{*}{ACM} & 0\% & / & 92.92$\pm$0.67& 92.85$\pm$0.68\\
    \multirow{3}{*}{} & 17\% &subject, term & 93.10$\pm$0.27& 93.14$\pm$0.26\\
    \multirow{3}{*}{} & 54\% &author, subject & 93.55$\pm$0.20& 93.47$\pm$0.21\\
    \multirow{3}{*}{} & 69\% & author, subject, term& \textbf{93.86$\pm$0.18} & \textbf{93.80$\pm$0.18} \\
    \midrule
    \multirow{3}{*}{IMDB} & 0\% & / & 62.98$\pm$1.66& 67.42$\pm$0.42\\
    \multirow{3}{*}{} & 37\% & keyword & 63.65$\pm$0.57& 67.52$\pm$0.36\\
    \multirow{3}{*}{} & 67\% &actor, keyword & 64.59$\pm$0.53& 67.86$\pm$0.42\\
    \multirow{3}{*}{} & 76\% & director, actor, keyword& \textbf{64.92$\pm$0.58} & \textbf{67.94$\pm$0.41}\\
    \bottomrule
  \end{tabular}}
\end{table}

\begin{table}[ht!]
  \caption{\textcolor{ black}{Performance of SimpleHGN-\autoac{} with varying masked edge rates in the link prediction task.}} 
  \centering
  \scalebox{0.80}{
  \label{tbl:edge rates}
  
  \begin{tabular}{cclcc}
    \toprule
    Datasets   & Masked Edge Rates &Models &ROC-AUC &MRR \\
    
    \midrule
    \multirow{8}{*}{DBLP} & \multirow{2}{*}{5\%} &SimpleHGN & 95.92$\pm$0.56 &	97.16$\pm$0.44\\
    \multirow{8}{*}{} &  \multirow{2}{*}{} &SimpleHGN-\autoac{} & \textbf{97.62$\pm$0.36} &	\textbf{99.02$\pm$0.24} \\
   \multirow{8}{*}{} & \multirow{2}{*}{10\%} &SimpleHGN & 94.61$\pm$0.11 &	97.21$\pm$0.16\\
    \multirow{8}{*}{} & \multirow{2}{*}{} &SimpleHGN-\autoac{} & \textbf{95.87$\pm$0.66} &	\textbf{98.21$\pm$0.21} \\
    \multirow{8}{*}{} & \multirow{2}{*}{20\%} &SimpleHGN & 91.34$\pm$0.61 &	95.65$\pm$0.41\\
    \multirow{8}{*}{} & \multirow{2}{*}{} &SimpleHGN-\autoac{} & \textbf{94.08$\pm$0.72} &	\textbf{97.61$\pm$0.33} \\
    \multirow{8}{*}{} & \multirow{2}{*}{30\%} &SimpleHGN & 88.76$\pm$0.66 &	95.39$\pm$0.24\\
    \multirow{8}{*}{} & \multirow{2}{*}{} &SimpleHGN-\autoac{} & \textbf{91.11$\pm$0.67} &	\textbf{97.42$\pm$0.44} \\
    \midrule
    \multirow{8}{*}{IMDB} & \multirow{2}{*}{5\%} &SimpleHGN & 64.89$\pm$0.58 &	81.86$\pm$0.94\\
    \multirow{8}{*}{} & \multirow{2}{*}{} &SimpleHGN-\autoac{} & \textbf{86.57$\pm$1.36} &	\textbf{92.75$\pm$0.84}\\
   \multirow{8}{*}{} & \multirow{2}{*}{10\%} &SimpleHGN & 57.92$\pm$2.32 &	79.09$\pm$1.40\\
    \multirow{8}{*}{} & \multirow{2}{*}{} &SimpleHGN-\autoac{} & \textbf{74.14$\pm$0.73} &	\textbf{86.27$\pm$0.45}\\
    \multirow{8}{*}{} & \multirow{2}{*}{20\%} &SimpleHGN & 58.21$\pm$0.39 &	79.71$\pm$0.34\\
    \multirow{8}{*}{} & \multirow{2}{*}{} &SimpleHGN-\autoac{} & \textbf{73.75$\pm$0.82} &	\textbf{86.25$\pm$0.32}\\
    \multirow{8}{*}{} & \multirow{2}{*}{30\%} &SimpleHGN & 54.13$\pm$0.79 &	77.57$\pm$0.67\\
    \multirow{8}{*}{} & \multirow{2}{*}{} &SimpleHGN-\autoac{} & \textbf{65.81$\pm$0.31} &	\textbf{83.23$\pm$0.21}\\
    \bottomrule
  \end{tabular}}
  \vspace{-2ex}
\end{table}

\subsection{\textcolor{ black}{Impacts of Attribute Missing Rates and Masked Edge Rates}}
\subsubsection{\textcolor{ black}{Study on the performance of the same dataset with varying attribute missing rates in the node classification task}}
\textcolor{ black}{
Table~\ref{tbl:attrmiss} shows the performance of SimpleHGN-\autoac{} with varying attribute missing rates.
We change attribute missing rates by completing the missing attributes with one-hot encoding, which is a commonly used handcrafted attribute completion method.
A missing rate of 0\% means that all missing attributes are completed manually. 
45\%, 69\%, and 76\% are inherent attribute missing rates of DBLP, ACM, and IMDB, respectively, i.e., only one node type has raw attributes. 
%
%
From Table~\ref{tbl:attrmiss}, we can see that SimpleHGN-\autoac{} performs better with higher missing rates, indicating that \autoac{} is capable of searching for the suitable completion operation for each no-attribute node and the searched completion operations are superior to the handcrafted completion method.
}

\subsubsection{\textcolor{ black}{Study on the performance of the same dataset with varying masked edge rates in the link prediction task}}
 \textcolor{ black}
{
%
Table~\ref{tbl:edge rates} shows the performance of SimpleHGN-\autoac{} with varying masked edge rates.
The edges are masked randomly.
We can see that SimpleHGN-\autoac{} achieves better performance than SimpleHGN at different masked edge rates, especially on the IMDB dataset.
Moreover, the performance of both models decreases as the masked edge rate increases.
}

\section{Conclusion}
In this paper, we proposed a differentiable attribute completion framework called \autoac{} for automated completion operation search in heterogeneous GNNs.
First, we introduced an expressive completion operation search space and proposed a continuous relaxation scheme to make the search space differentiable.
Second, we formulated the completion operation search as a bi-level joint optimization problem.
%
To improve search efficiency, we enforced discrete constraints
on completion parameters and further proposed a proximal iteration-based search algorithm. 
Moreover, we leveraged an auxiliary unsupervised node clustering task to reduce the dimension of completion parameters.
Extensive experimental results reveal that \autoac{} is effective to boost the performance of heterogeneous GNNs and outperforms the SOTA attribute completion method in terms of performance and efficiency.


\section*{Acknowledgment}
This work was supported by the National Natural Science Foundation of China (\#62102177), the Natural Science Foundation of Jiangsu Province (\#BK20210181), the Key R\&D Program of Jiangsu Province (\#BE2021729), Open Research Projects of Zhejiang Lab (\#2022PG0AB07), and the Collaborative Innovation Center of Novel Software Technology and Industrialization, Jiangsu, China.
Guanghui Zhu and Yihua Huang are corresponding authors with equal contributions.

\bibliographystyle{IEEEtran}
\bibliography{reference}

\begin{thebibliography}{10}
\providecommand{\url}[1]{#1}
\csname url@samestyle\endcsname
\providecommand{\newblock}{\relax}
\providecommand{\bibinfo}[2]{#2}
\providecommand{\BIBentrySTDinterwordspacing}{\spaceskip=0pt\relax}
\providecommand{\BIBentryALTinterwordstretchfactor}{4}
\providecommand{\BIBentryALTinterwordspacing}{\spaceskip=\fontdimen2\font plus
\BIBentryALTinterwordstretchfactor\fontdimen3\font minus
  \fontdimen4\font\relax}
\providecommand{\BIBforeignlanguage}[2]{{%
\expandafter\ifx\csname l@#1\endcsname\relax
\typeout{** WARNING: IEEEtran.bst: No hyphenation pattern has been}%
\typeout{** loaded for the language `#1'. Using the pattern for}%
\typeout{** the default language instead.}%
\else
\language=\csname l@#1\endcsname
\fi
#2}}
\providecommand{\BIBdecl}{\relax}
\BIBdecl

\bibitem{hamilton2017inductive}
W.~Hamilton, Z.~Ying, and J.~Leskovec, ``Inductive representation learning on
  large graphs,'' \emph{Advances in neural information processing systems},
  vol.~30, 2017.

\bibitem{sen2008collective}
P.~Sen, G.~Namata, M.~Bilgic, L.~Getoor, B.~Galligher, and T.~Eliassi-Rad,
  ``Collective classification in network data,'' \emph{AI magazine}, vol.~29,
  no.~3, pp. 93--93, 2008.

\bibitem{zitnik2017predicting}
M.~Zitnik and J.~Leskovec, ``Predicting multicellular function through
  multi-layer tissue networks,'' \emph{Bioinformatics}, vol.~33, no.~14, pp.
  i190--i198, 2017.

\bibitem{ji2021survey}
S.~Ji, S.~Pan, E.~Cambria, P.~Marttinen, and S.~Y. Philip, ``A survey on
  knowledge graphs: Representation, acquisition, and applications,'' \emph{IEEE
  Transactions on Neural Networks and Learning Systems}, vol.~33, no.~2, pp.
  494--514, 2021.

\bibitem{sun2013mining}
Y.~Sun and J.~Han, ``Mining heterogeneous information networks: a structural
  analysis approach,'' \emph{Acm Sigkdd Explorations Newsletter}, vol.~14,
  no.~2, pp. 20--28, 2013.

\bibitem{shi2016survey}
C.~Shi, Y.~Li, J.~Zhang, Y.~Sun, and S.~Y. Philip, ``A survey of heterogeneous
  information network analysis,'' \emph{IEEE Transactions on Knowledge and Data
  Engineering}, vol.~29, no.~1, pp. 17--37, 2016.

\bibitem{wu2020comprehensive}
Z.~Wu, S.~Pan, F.~Chen, G.~Long, C.~Zhang, and S.~Y. Philip, ``A comprehensive
  survey on graph neural networks,'' \emph{IEEE transactions on neural networks
  and learning systems}, vol.~32, no.~1, pp. 4--24, 2020.

\bibitem{kipf2017gcn}
T.~N. Kipf and M.~Welling, ``Semi-supervised classification with graph
  convolutional networks,'' in \emph{International Conference on Learning
  Representations}, 2017.

\bibitem{medical2021sigmod}
\BIBentryALTinterwordspacing
A.~Vretinaris, C.~Lei, V.~Efthymiou, X.~Qin, and F.~\"{O}zcan, ``Medical entity
  disambiguation using graph neural networks,'' in \emph{Proceedings of the
  2021 International Conference on Management of Data}, ser. SIGMOD '21.\hskip
  1em plus 0.5em minus 0.4em\relax New York, NY, USA: Association for Computing
  Machinery, 2021, p. 2310–2318. [Online]. Available:
  \url{https://doi.org/10.1145/3448016.3457328}
\BIBentrySTDinterwordspacing

\bibitem{wang2019heterogeneous}
X.~Wang, H.~Ji, C.~Shi, B.~Wang, Y.~Ye, P.~Cui, and P.~S. Yu, ``Heterogeneous
  graph attention network,'' in \emph{The world wide web conference}, 2019, pp.
  2022--2032.

\bibitem{yun2019graph}
S.~Yun, M.~Jeong, R.~Kim, J.~Kang, and H.~J. Kim, ``Graph transformer
  networks,'' \emph{Advances in neural information processing systems},
  vol.~32, 2019.

\bibitem{zhu2019relation}
S.~Zhu, C.~Zhou, S.~Pan, X.~Zhu, and B.~Wang, ``Relation structure-aware
  heterogeneous graph neural network,'' in \emph{2019 IEEE international
  conference on data mining (ICDM)}.\hskip 1em plus 0.5em minus 0.4em\relax
  IEEE, 2019, pp. 1534--1539.

\bibitem{zhang2019heterogeneous}
C.~Zhang, D.~Song, C.~Huang, A.~Swami, and N.~V. Chawla, ``Heterogeneous graph
  neural network,'' in \emph{Proceedings of the 25th ACM SIGKDD international
  conference on knowledge discovery \& data mining}, 2019, pp. 793--803.

\bibitem{fu2020magnn}
X.~Fu, J.~Zhang, Z.~Meng, and I.~King, ``Magnn: Metapath aggregated graph
  neural network for heterogeneous graph embedding,'' in \emph{Proceedings of
  The Web Conference 2020}, 2020, pp. 2331--2341.

\bibitem{hu2020heterogeneous}
Z.~Hu, Y.~Dong, K.~Wang, and Y.~Sun, ``Heterogeneous graph transformer,'' in
  \emph{Proceedings of The Web Conference 2020}, 2020, pp. 2704--2710.

\bibitem{hong2020attention}
H.~Hong, H.~Guo, Y.~Lin, X.~Yang, Z.~Li, and J.~Ye, ``An attention-based graph
  neural network for heterogeneous structural learning,'' in \emph{Proceedings
  of the AAAI conference on artificial intelligence}, vol.~34, no.~04, 2020,
  pp. 4132--4139.

\bibitem{lv2021we}
Q.~Lv, M.~Ding, Q.~Liu, Y.~Chen, W.~Feng, S.~He, C.~Zhou, J.~Jiang, Y.~Dong,
  and J.~Tang, ``Are we really making much progress? revisiting, benchmarking
  and refining heterogeneous graph neural networks,'' in \emph{Proceedings of
  the 27th ACM SIGKDD Conference on Knowledge Discovery \& Data Mining}, 2021,
  pp. 1150--1160.

\bibitem{jin2021heterogeneous}
D.~Jin, C.~Huo, C.~Liang, and L.~Yang, ``Heterogeneous graph neural network via
  attribute completion,'' in \emph{Proceedings of the Web Conference 2021},
  2021, pp. 391--400.

\bibitem{yao2020efficient}
Q.~Yao, J.~Xu, W.-W. Tu, and Z.~Zhu, ``Efficient neural architecture search via
  proximal iterations,'' in \emph{Proceedings of the AAAI Conference on
  Artificial Intelligence}, vol.~34, no.~04, 2020, pp. 6664--6671.

\bibitem{velickovic2017graph}
P.~Velickovic, G.~Cucurull, A.~Casanova, A.~Romero, P.~Lio, and Y.~Bengio,
  ``Graph attention networks,'' \emph{stat}, vol. 1050, p.~20, 2017.

\bibitem{10.1145/3514221.3517872}
\BIBentryALTinterwordspacing
D.~Yao, Y.~Gu, G.~Cong, H.~Jin, and X.~Lv, ``Entity resolution with
  hierarchical graph attention networks,'' in \emph{Proceedings of the 2022
  International Conference on Management of Data}, ser. SIGMOD '22.\hskip 1em
  plus 0.5em minus 0.4em\relax New York, NY, USA: Association for Computing
  Machinery, 2022, p. 429–442. [Online]. Available:
  \url{https://doi.org/10.1145/3514221.3517872}
\BIBentrySTDinterwordspacing

\bibitem{10.1145/3318464.3389706}
\BIBentryALTinterwordspacing
W.~Zhang, X.~Miao, Y.~Shao, J.~Jiang, L.~Chen, O.~Ruas, and B.~Cui, ``Reliable
  data distillation on graph convolutional network,'' in \emph{Proceedings of
  the 2020 ACM SIGMOD International Conference on Management of Data}, ser.
  SIGMOD '20.\hskip 1em plus 0.5em minus 0.4em\relax New York, NY, USA:
  Association for Computing Machinery, 2020, p. 1399–1414. [Online].
  Available: \url{https://doi.org/10.1145/3318464.3389706}
\BIBentrySTDinterwordspacing

\bibitem{sun2018rotate}
\BIBentryALTinterwordspacing
Z.~Sun, Z.-H. Deng, J.-Y. Nie, and J.~Tang, ``Rotate: Knowledge graph embedding
  by relational rotation in complex space,'' in \emph{International Conference
  on Learning Representations}, 2019. [Online]. Available:
  \url{https://openreview.net/forum?id=HkgEQnRqYQ}
\BIBentrySTDinterwordspacing

\bibitem{yu2021gcn}
Z.~Yu, D.~Jin, Z.~Liu, D.~He, X.~Wang, H.~Tong, and J.~Han, ``As-gcn: Adaptive
  semantic architecture of graph convolutional networks for text-rich
  networks,'' in \emph{IEEE International Conference on Data Mining
  (ICDM)}.\hskip 1em plus 0.5em minus 0.4em\relax IEEE, 2021, pp. 837--846.

\bibitem{dong2017metapath2vec}
Y.~Dong, N.~V. Chawla, and A.~Swami, ``metapath2vec: Scalable representation
  learning for heterogeneous networks,'' in \emph{Proceedings of the 23rd ACM
  SIGKDD international conference on knowledge discovery and data mining},
  2017, pp. 135--144.

\bibitem{9724614}
D.~He, C.~Liang, C.~Huo, Z.~Feng, D.~Jin, L.~Yang, and W.~Zhang, ``Analyzing
  heterogeneous networks with missing attributes by unsupervised contrastive
  learning,'' \emph{IEEE Transactions on Neural Networks and Learning Systems},
  pp. 1--13, 2022.

\bibitem{bayram2021node}
E.~Bayram, A.~Garc{\'\i}a-Dur{\'a}n, and R.~West, ``Node attribute completion
  in knowledge graphs with multi-relational propagation,'' in \emph{IEEE
  International Conference on Acoustics, Speech and Signal Processing
  (ICASSP)}.\hskip 1em plus 0.5em minus 0.4em\relax IEEE, 2021, pp. 3590--3594.

\bibitem{elsken2019neural}
T.~Elsken, J.~H. Metzen, and F.~Hutter, ``Neural architecture search: A
  survey,'' \emph{The Journal of Machine Learning Research}, vol.~20, no.~1,
  pp. 1997--2017, 2019.

\bibitem{wang2022automated}
X.~Wang, Z.~Zhang, and W.~Zhu, ``Automated graph machine learning: Approaches,
  libraries and directions,'' \emph{arXiv preprint arXiv:2201.01288}, 2022.

\bibitem{lai2020policy}
K.-H. Lai, D.~Zha, K.~Zhou, and X.~Hu, ``Policy-gnn: Aggregation optimization
  for graph neural networks,'' in \emph{Proceedings of the 26th ACM SIGKDD
  International Conference on Knowledge Discovery \& Data Mining}, 2020, pp.
  461--471.

\bibitem{zhao2020simplifying}
H.~Zhao, L.~Wei, and Q.~Yao, ``Simplifying architecture search for graph neural
  network,'' \emph{arXiv preprint arXiv:2008.11652}, 2020.

\bibitem{gao2019graphnas}
Y.~Gao, H.~Yang, P.~Zhang, C.~Zhou, and Y.~Hu, ``Graphnas: Graph neural
  architecture search with reinforcement learning,'' \emph{arXiv preprint
  arXiv:1904.09981}, 2019.

\bibitem{zhou2019auto}
K.~Zhou, Q.~Song, X.~Huang, and X.~Hu, ``Auto-gnn: Neural architecture search
  of graph neural networks,'' \emph{arXiv preprint arXiv:1909.03184}, 2019.

\bibitem{zhu2022psp}
G.~Zhu, W.~Wang, Z.~Xu, F.~Cheng, M.~Qiu, C.~Yuan, and Y.~Huang, ``Psp:
  Progressive space pruning for efficient graph neural architecture search,''
  in \emph{IEEE 38th International Conference on Data Engineering
  (ICDE)}.\hskip 1em plus 0.5em minus 0.4em\relax IEEE, 2022, pp. 2168--2181.

\bibitem{han2020genetic}
Z.~Han, F.~Xu, J.~Shi, Y.~Shang, H.~Ma, P.~Hui, and Y.~Li, ``Genetic
  meta-structure search for recommendation on heterogeneous information
  network,'' in \emph{Proceedings of the 29th ACM International Conference on
  Information \& Knowledge Management}, 2020, pp. 455--464.

\bibitem{ding2021diffmg}
Y.~Ding, Q.~Yao, H.~Zhao, and T.~Zhang, ``Diffmg: Differentiable meta graph
  search for heterogeneous graph neural networks,'' in \emph{Proceedings of the
  27th ACM SIGKDD Conference on Knowledge Discovery \& Data Mining}, 2021, pp.
  279--288.

\bibitem{parikh2014proximal}
N.~Parikh, S.~Boyd \emph{et~al.}, ``Proximal algorithms,'' \emph{Foundations
  and trends{\textregistered} in Optimization}, vol.~1, no.~3, pp. 127--239,
  2014.

\bibitem{klicpera2018predict}
J.~Klicpera, A.~Bojchevski, and S.~G{\"u}nnemann, ``Predict then propagate:
  Graph neural networks meet personalized pagerank,'' in \emph{International
  Conference on Learning Representations}, 2019.

\bibitem{finn2017maml}
C.~Finn, P.~Abbeel, and S.~Levine, ``Model-agnostic meta-learning for fast
  adaptation of deep networks,'' in \emph{Proceedings of the 34th International
  Conference on Machine Learning-Volume 70}, 2017, pp. 1126--1135.

\bibitem{liu2018darts}
H.~Liu, K.~Simonyan, and Y.~Yang, ``Darts: Differentiable architecture
  search,'' in \emph{International Conference on Learning Representations},
  2018.

\bibitem{dempster1977maximum}
A.~P. Dempster, N.~M. Laird, and D.~B. Rubin, ``Maximum likelihood from
  incomplete data via the em algorithm,'' \emph{Journal of the Royal
  Statistical Society: Series B (Methodological)}, vol.~39, no.~1, pp. 1--22,
  1977.

\bibitem{macqueen1967classification}
J.~MacQueen, ``Classification and analysis of multivariate observations,'' in
  \emph{5th Berkeley Symp. Math. Statist. Probability}, 1967, pp. 281--297.

\bibitem{good2010performance}
B.~H. Good, Y.-A. De~Montjoye, and A.~Clauset, ``Performance of modularity
  maximization in practical contexts,'' \emph{Physical review E}, vol.~81,
  no.~4, p. 046106, 2010.

\bibitem{bianchi2020spectral}
F.~M. Bianchi, D.~Grattarola, and C.~Alippi, ``Spectral clustering with graph
  neural networks for graph pooling,'' in \emph{International Conference on
  Machine Learning}.\hskip 1em plus 0.5em minus 0.4em\relax PMLR, 2020, pp.
  874--883.

\bibitem{tsitsulin2020graph}
A.~Tsitsulin, J.~Palowitch, B.~Perozzi, and E.~M{\"u}ller, ``Graph clustering
  with graph neural networks,'' \emph{arXiv preprint arXiv:2006.16904}, 2020.

\bibitem{kingma2014adam}
D.~P. Kingma and J.~Ba, ``Adam: A method for stochastic optimization,''
  \emph{arXiv preprint arXiv:1412.6980}, 2014.

\bibitem{cen2019representation}
Y.~Cen, X.~Zou, J.~Zhang, H.~Yang, J.~Zhou, and J.~Tang, ``Representation
  learning for attributed multiplex heterogeneous network,'' in
  \emph{Proceedings of the 25th ACM SIGKDD International Conference on
  Knowledge Discovery \& Data Mining}, 2019, pp. 1358--1368.

\bibitem{cantador2011second}
I.~Cantador, P.~Brusilovsky, and T.~Kuflik, ``Second workshop on information
  heterogeneity and fusion in recommender systems (hetrec2011),'' in
  \emph{Proceedings of the fifth ACM conference on Recommender systems}, 2011,
  pp. 387--388.

\end{thebibliography}
\clearpage
\appendix

\subsection{Details of Datasets}
\label{ss:details}
DBLP\footnote{ https://dblp.uni-trier.de/} is a computer science bibliography website. 
%
The raw attribute of the paper node is the bag-of-words representation of keywords. 
%
ACM\footnote{ http://dl.acm.org/} is a citation network.
%
%
The raw attribute of the paper node is also the bag-of-words representation of keywords.
%
IMDB\footnote{ https://www.imdb.com} is a website about movies. 
The attributes of movie nodes are originally present, they are represented by the bag-of-words representation of words extracted for key episodes of movies.
%
%
LastFM is extracted from last.fm with timestamps from January 2015 to June 2015.
We use the subset released by~\cite{cantador2011second}.
The target is to predict whether a user likes a certain artist.
The raw attribute of the artist node is the one-hot encoding. 
For the DBLP dataset, the attributes of the target nodes are missing.
For the ACM and IMDB datasets, the target nodes have raw attributes.

\subsection{Implementations and Configurations of Baselines}
\label{ss:conf}
We use the HGB benchmark to evaluate the performance of all baselines. 
In HGB, implementations of baselines are based on their official codes to avoid errors introduced by re-implementation.
Next, we present the configurations of baselines in the node classification and link prediction tasks, respectively.
For brevity, we denote the dimension of node embedding as $d$, the dimension of edge embedding as $d_e$, the dimension of attention vector (if exists) as $d_a$, the number of GNN layers as $L$, the number of attention heads as $n_h$, the negative slope of LeakyReLU as $s$.
\subsubsection{Node Classification}
The baselines in the node classification task contain HAN, GTN, HetSANN, MAGNN, HGCA, HGT, HetGNN, GCN, GAT, SimpleHGN, and HGNN-AC.
\begin{itemize}
\item \emph{HAN:} We set $d=8$, $d_a=128$, $n_h=8$, and $L=2$ for all datasets. 
\item \emph{GTN:} The adaptive learning rate is employed for all datasets. We set $d=64$ and the number of GTN channels to 2. For DBLP and ACM, we set $L=2$. For IMDB, we set $L=3$. 
\item \emph{HetSANN:} For ACM, we set $d=64$, $L=3$, and $n_h=8$. For IMDB, we set $d=32$, $L=2$, and $n_h=4$. For DBLP, we set $d=64$, $L=2$, and $n_h=4$.
\item \emph{MAGNN:} For DBLP and ACM, we set the batch size to 8, and the number of neighbor samples to 100. For IMDB, we use full batch training.
\item \emph{HGCA:} We set $d=64$, the temperature parameter $\tau$ = 0.5, and the loss coefficient $\lambda$ = 0.5.
\item \emph{HGT:} We use the layer normalization in each layer, and set $d=64$ and $n_h=8$ for all datasets.
$L$ is set to 2, 3, 5 for ACM, DBLP and IMDB,respectively.
\item \emph{HetGNN:} We set $d=128$, and the batch size to 200 for all datasets. For random walk, we set the walk length to 30 and the window size to 5.
\item \emph{GCN:} We set $d=64$ for all datasets. We set $L=3$ for DBLP and ACM, and $L=4$ for IMDB. 
\item \emph{GAT:} We set $d=64$ and $n_h=8$ for all datasets. For DBLP and ACM, we set $s=0.05$ and $L=3$. For IMDB, we set $s=0.1$ and $L=5$. 
\item \emph{SimpleHGN:} We set $d=d_e=64$, $n_h=8$, and the edge residual $\beta=0.05$ for all datasets. For DBLP and ACM, we set $L=3$ and $s=0.05$. For IMDB, we set $L=6$ and $s=0.1$. 
\item \emph{HGNN-AC:} We set $d=64$, $n_h=8$, the divided ratio $\alpha$ of $N^+$ to 0.3, and the loss weighted coefficient $\lambda$ to 0.5 for all datasets, which are consistent with the original paper.
\end{itemize}

\subsubsection{Link prediction}
The baselines in the link prediction task contain GATNE, HetGNN, GCN, GAT, and SimpleHGN.
\begin{itemize}
  
    \item \emph{GATNE:} We set $d=200$, $d_e=10$, and $d_a=20$ for all datasets.
    For the random walk, we set the walk length to 30 and the window size to 5. For neighbor sampling, we set the number of negative samples for optimization to 5 and the number of neighbor samples for aggregation to 10.
    \item \emph{HetGNN:} We set $d=128$, and the batch size to 200 for all datasets. For random walk, we set the walk length to 30 and the window size to 5.
    \item \emph{GCN:} We set $d=64$ and $L=2$ for all datasets.
    \item \emph{GAT:} For LastFM, we set $d=64$, $n_h=4$, $L=3$, and $s=0.1$. For DBLP, we set $d=64$, $n_h=8$, $L=3$, and $s=0.05$. For IMDB, we set $d=64$, $n_h=4$, $L=5$, and $s=0.1$.
    \item \emph{SimpleHGN:} We set $d=64$, $d_e=32$, $n_h=2$, the edge residual $\beta=0$, and $s=0.01$ for all datasets. For DBLP, we set $L=3$. For LastFM, we set $L=4$. For IMDB, we set $L=6$.
    
\end{itemize}

\begin{figure}[t]
  \centering
    \subfigure[DBLP]{
    \includegraphics[width=0.29\columnwidth]{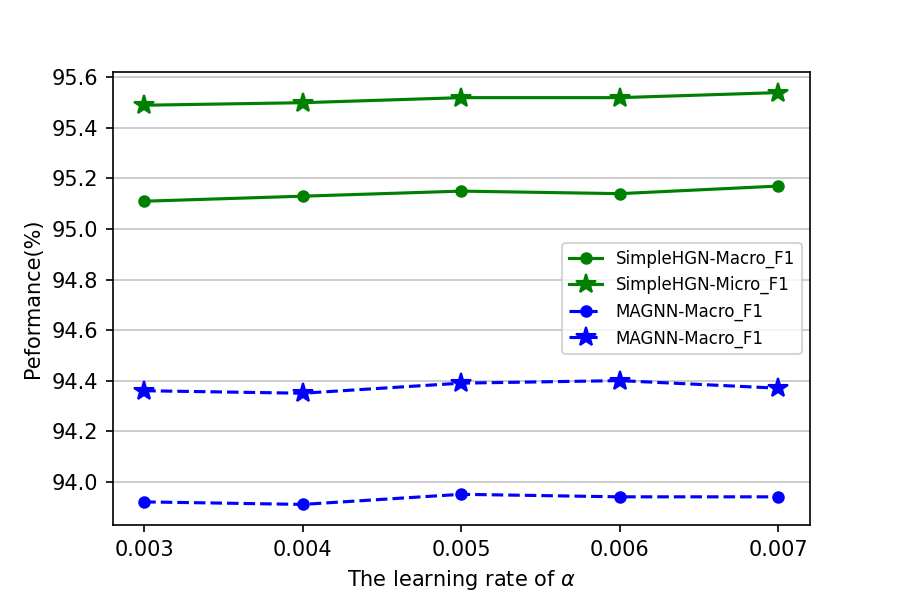}
    }
    \subfigure[ACM]{
    \includegraphics[width=0.29\columnwidth]{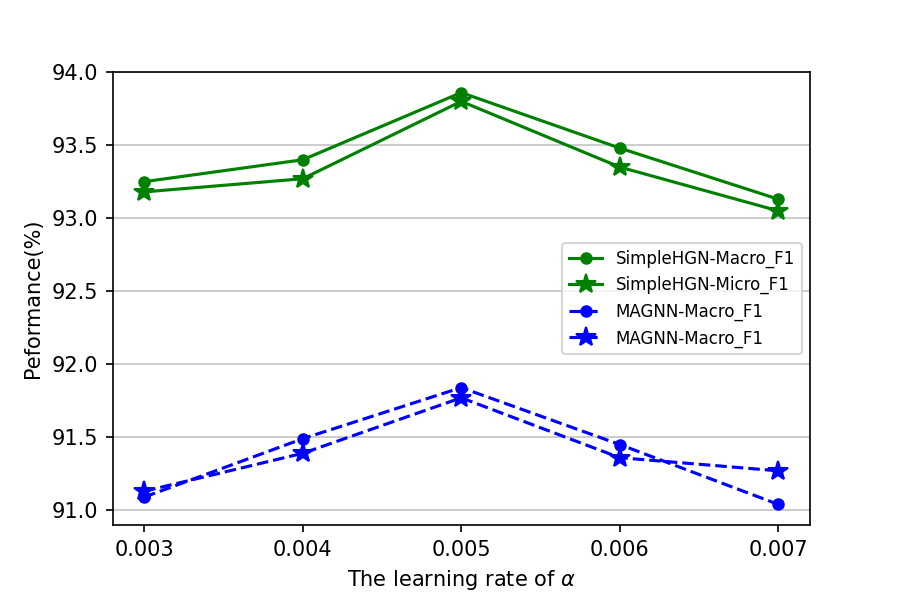}
    }
    \subfigure[IMDB]{
    \includegraphics[width=0.29\columnwidth]{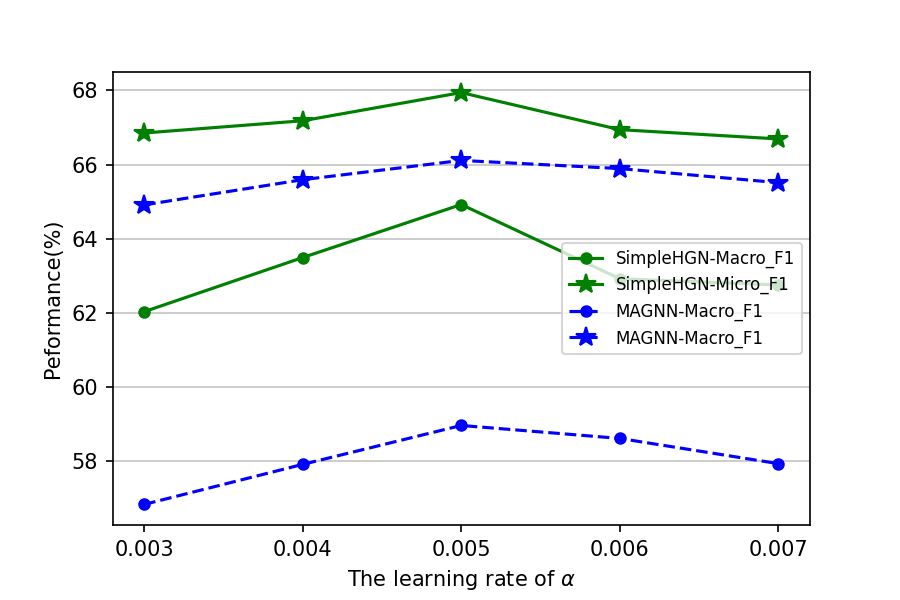}
    }

  \caption{Performance comparison under different learning rates}
  \label{Performance comparison of AutoHC under different lr}
  \vspace{-1ex}
\end{figure}
\begin{figure}[t]
  \centering
   
    \subfigure[DBLP]{
    \includegraphics[width=0.29\columnwidth]{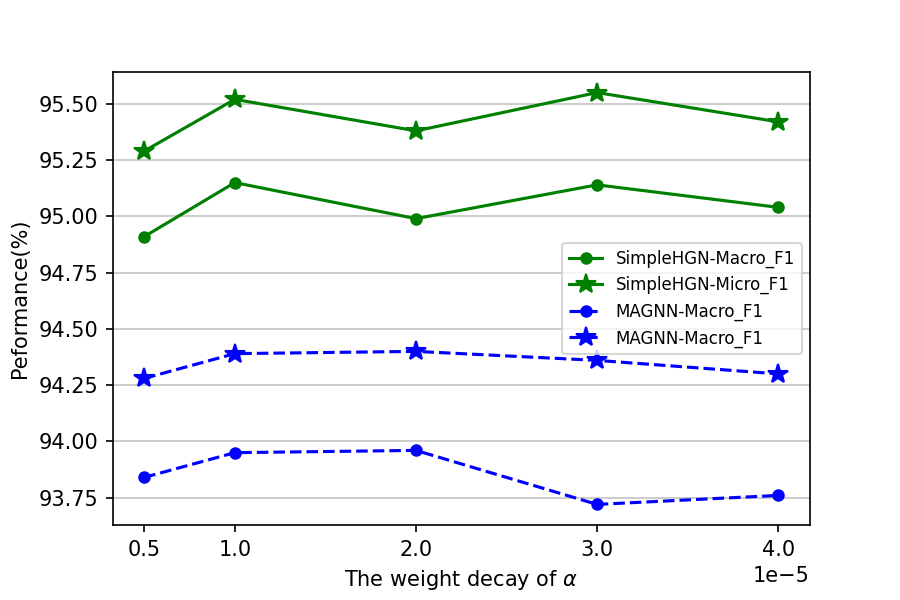}
    }
    \subfigure[ACM]{
    \includegraphics[width=0.29\columnwidth]{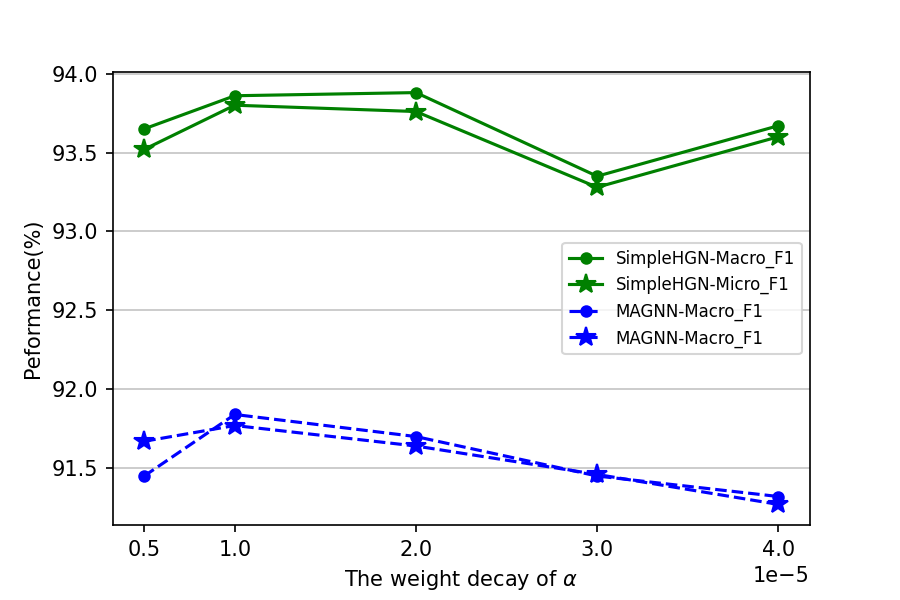}
    }
    \subfigure[IMDB]{
    \includegraphics[width=0.29\columnwidth]{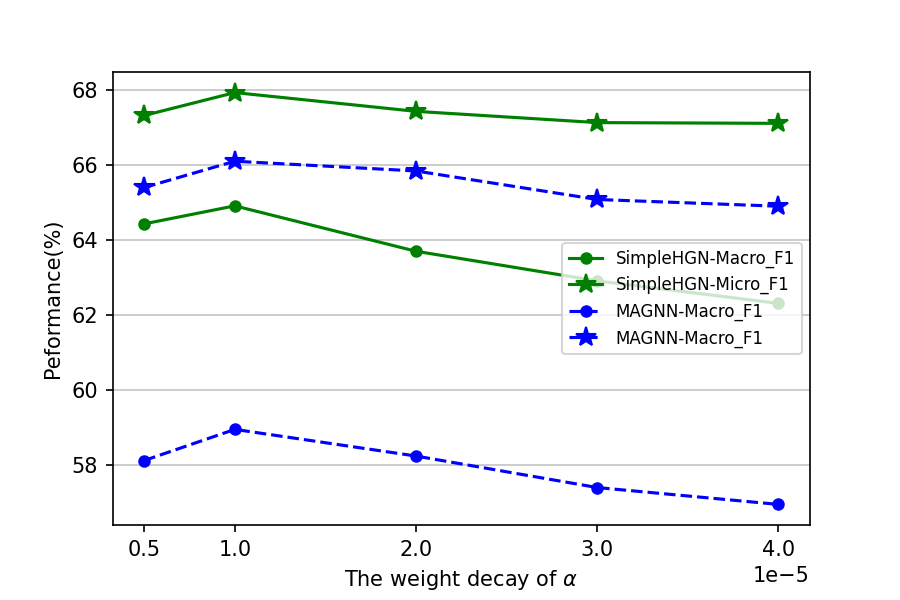}
    }

  \caption{Performance comparison under different weight decay values}
  \label{Performance comparison of AutoHC under different wd}
  \vspace{-1ex}
\end{figure}

\subsection{Effects of the learning rate and the weight decay}
\label{ss:lr}
We further evaluate the effect of the learning rate and weight decay when optimizing the completion parameters $\alpha$.
The available learning rates are set to [3e-3, 4e-3, 5e-3, 6e-3, 7e-3].
The available weight decay values are set to [5e-6,1e-5, 2e-5, 3e-5, 4e-3].
Figure~\ref{Performance comparison of AutoHC under different lr} and Figure~\ref{Performance comparison of AutoHC under different wd} show the performances of \autoac{} with different learning rates and different weight decay, respectively.
The green and blue lines represent SimpleHGN-\autoac{} and MAGNN-\autoac{}, respectively.
From Figure~\ref{Performance comparison of AutoHC under different lr} and Figure~\ref{Performance comparison of AutoHC under different wd}, we can see that \autoac{} is very robust to the learning rate and the weight decay.

\end{document}